\documentclass[10pt,twocolumn,letterpaper]{article}

\usepackage[pagenumbers]{cvpr} 

\usepackage{amsmath,amsfonts,bm}

\def\eqref#1{equation~\ref{#1}}

\def\1{\bm{1}}

\DeclareMathAlphabet{\mathsfit}{\encodingdefault}{\sfdefault}{m}{sl}
\SetMathAlphabet{\mathsfit}{bold}{\encodingdefault}{\sfdefault}{bx}{n}

\usepackage[T1]{fontenc}
\usepackage{amsfonts}
\usepackage{nicefrac}
\usepackage{microtype}
\usepackage{colortbl}
\usepackage{wrapfig}
\usepackage{enumitem}
\usepackage{algorithm}
\usepackage{algorithmic}
\usepackage[most]{tcolorbox}
\usepackage{listings}
\usepackage{fvextra}
\RecustomVerbatimEnvironment{verbatim}{Verbatim}{breaklines=true,breakanywhere=true,fontsize=\small}
\usepackage{adjustbox}
\usepackage{array}
\usepackage{tabularx}
\newcolumntype{Y}{>{\centering\arraybackslash}X}
\usepackage{multirow}
\usepackage{pifont}

\newtcolorbox{PromptBox}[1]{
    enhanced,
    colback=gray!5,
    colframe=black,
    fonttitle=\bfseries,
    colbacktitle=white,
    coltitle=black,
    attach boxed title to top left={yshift=-3mm, xshift=3mm},
    boxed title style={
        colframe=black,
        arc=1.5mm,
        size=small
    },
    title={#1},
    arc=3mm,
    left=10pt, right=10pt, top=12pt, bottom=10pt,
    boxrule=0.8pt
}

\definecolor{codeboxaccent}{HTML}{888888}
\newtcblisting[auto counter]{codebox}[3][]{
    enhanced,
    listing only,
    listing engine=listings,
    listing options={
        language=#2,
        basicstyle=\ttfamily\footnotesize,
        keywordstyle=\bfseries,
        numbers=none,
        breaklines=false,
        columns=fullflexible,
        keepspaces=true,
        showstringspaces=false
    },
    colback=gray!3,
    colframe=codeboxaccent,
    colbacktitle=codeboxaccent,
    coltitle=white,
    title={Code~\thetcbcounter\quad #3},
    fonttitle=\bfseries\small,
    boxrule=0.6pt,
    arc=2mm,
    left=4mm,
    lefttitle=0.5mm,
    right=1.5mm,
    top=1.5mm,
    bottom=1.5mm,
    #1
}

\newcommand{\cmark}{\ding{51}}
\newcommand{\xmark}{\ding{55}}
\definecolor{our_color}{HTML}{666666}

\definecolor{cvprblue}{rgb}{0.21,0.49,0.74}
\usepackage[pagebackref,breaklinks,colorlinks,allcolors=cvprblue]{hyperref}
\usepackage{xr-hyper}
\def\confName{CVPR}
\def\confYear{2027}

\title{StreamPro: Reinforcement Learning with Multi-Grained Rewards \\ for Proactive Streaming Video Understanding}

\author{
Ao Li\thanks{Equal contribution to this work.}\textsuperscript{\rm ~~,1,2} \quad 
Zihan Xiao\footnotemark[1]\textsuperscript{\rm ~~,1,2} \quad 
Zihao Yue\textsuperscript{1} \quad
Boshen Xu\textsuperscript{1} \quad
Linli Yao\textsuperscript{3} \\
Jiaze Li\textsuperscript{2} \quad
Pei Fu\textsuperscript{2} \quad
Jianzhong Ju\textsuperscript{2} \quad
Jian Luan\textsuperscript{2} \quad
Qin Jin\thanks{Corresponding author.}\textsuperscript{\rm ~~,1}
\\
\textsuperscript{1}AIM3 Lab, Renmin University of China\quad
\textsuperscript{2}MiLM Plus, Xiaomi Inc.\quad
\textsuperscript{3}Peking University
\\
\tt{liaolea0808@gmail.com}
}

\begin{document}
\maketitle
\begin{abstract}
Proactive streaming video understanding requires models to continuously monitor evolving video streams and decide whether, when, and what to respond. 
However, training models with such proactive capabilities remains challenging. 
Supervised fine-tuning (SFT) suffers from severe imbalance between silence and response signals, while reinforcement learning (RL) faces sparse response-level rewards and difficulties in optimizing interdependent response sequences. 
To address these challenges, we propose \textbf{StreamPro}, a two-stage training framework. 
We first introduce \textbf{CB-Stream Loss} to alleviate supervision imbalance during SFT. 
We then employ Group Relative Policy Optimization (GRPO) with two complementary rewards: a refined \textbf{Response-Matching (RM)} reward that provides denser feedback for individual response decisions, and a rubric-based \textbf{Sequence-Quality (SQ)} reward that encourages globally coherent response sequences. 
To support the two training stages, we construct StreamPro-SFT-63K and StreamPro-RL-3K, respectively. 
To address the limited task coverage of existing proactive benchmarks, we further introduce \textbf{StreamPro-Bench}, encompassing three task categories: Perceptual Understanding, Temporal Reasoning, and Anticipatory Assistance, along with StreamPro-F1 to evaluate response correctness, timing, and frequency. 
Experiments demonstrate that StreamPro substantially improves proactive streaming video understanding, achieving an average StreamPro-F1 of 32.8 on StreamPro-Bench, compared with 9.2 for the strongest baseline, while maintaining strong performance on real-time streaming tasks, reaching 79.8 on StreamingBench-RTVU.

\end{abstract}

\section{Introduction}
\label{sec:introduction}

\begin{figure*}[tb]
    \centering
    \includegraphics[width=0.8\linewidth]{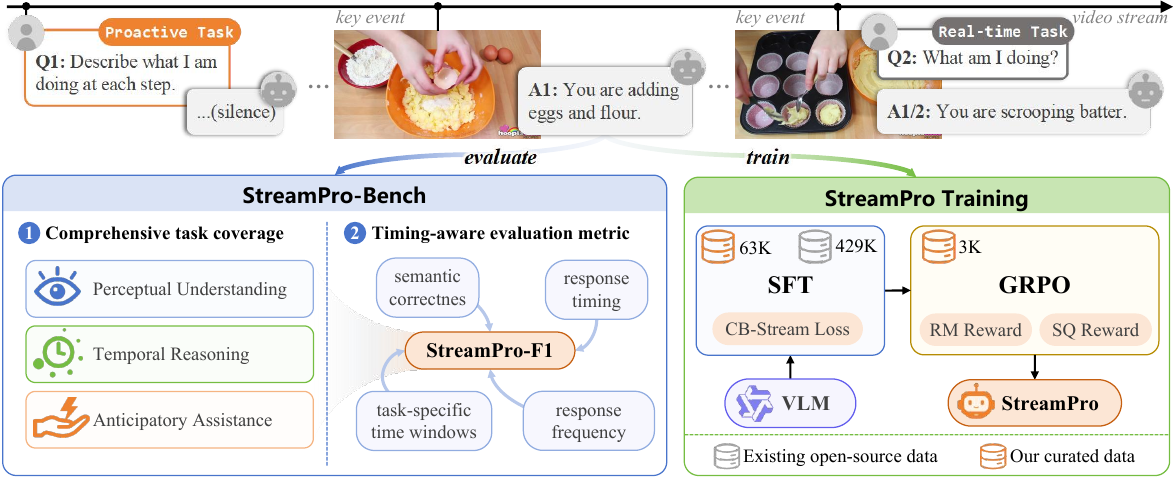}
    \caption{
        StreamPro Overview.
        \textbf{Top:} Difference between proactive and real-time tasks.
        \textbf{Bottom left:} StreamPro-Bench is a comprehensive benchmark covering three dimensions of proactive capability. Its StreamPro-F1 metric jointly assesses response timing and semantic correctness, uses task-specific time windows, and penalizes excessive and missed responses.
        \textbf{Bottom right:} StreamPro's two-stage training consists of SFT with CB-Stream Loss and GRPO with RM and SQ rewards.
        }
    \label{fig:teaser}
\end{figure*}

Streaming video understanding~\citep{videollmonline} requires models to continuously process visual inputs and make decisions as a video unfolds.
In conventional streaming settings, models primarily focus on understanding the current context or responding to user queries. 
In proactive streaming video understanding setting~\citep{mmduet,mmduet2,streamo}, where the events of interest occur after the user poses a query, models must determine not only \emph{what} to say, but also \emph{whether} and \emph{when} to respond.
Such capabilities are essential for real-world assistants to provide timely and context-aware assistance.

Despite recent progress, developing effective proactive streaming video understanding models remains challenging, both in learning when and what to respond and in comprehensively evaluating proactive capabilities.
\textit{\textbf{(1) Challenges in training proactive models.}}
During supervised fine-tuning (SFT), streaming data typically contains far more time steps requiring silence than those requiring responses. 
This imbalance causes training signals to be dominated by silence, making it difficult for models to learn when to respond~\citep{streamo,aura}.
During reinforcement learning (RL), rewards that strictly require both correct response content and precise timing often lead to sparse feedback, providing limited guidance for responses that are partially correct or slightly mistimed.
Furthermore, proactive tasks often involve multiple interdependent responses. Even if individual responses are reasonable, the overall response sequence may still suffer from redundancy, omissions, poor timing, or inconsistencies.
Therefore, effective training requires both fine-grained response-level feedback and holistic sequence-level guidance.
\textit{\textbf{(2) Limited task coverage in evaluation.}}
Existing benchmarks cover only a limited range of proactive tasks, making it difficult to comprehensively evaluate proactive capabilities~\citep{ovobench,streamingbench}, as shown in Table~\ref{tab:comparison_bench}.
Beyond basic perceptual understanding, proactive models should also be capable of reasoning about how events evolve over time and anticipating potential future situations.
For example, models should be able to recognize an impending danger and alert the user before it occurs.

To address these challenges, we introduce \textbf{StreamPro}, a two-stage training framework for proactive streaming video understanding. 
As illustrated in Fig.~\ref{fig:teaser}, our framework consists of SFT and RL stages. 
In the SFT stage, we introduce \textit{CB-Stream Loss}, which applies class-balanced reweighting~\citep{cb} to alleviate the severe imbalance between silence and response supervision. 
In the RL stage, we adopt GRPO~\citep{grpo} with two complementary rewards. 
The \emph{Response-Matching (RM)} reward is refined to provide denser feedback on response timing and semantic correctness. 
The rubric-based \emph{Sequence-Quality (SQ)} reward evaluates the quality of the complete response sequence, including its granularity, sequencing, and coverage. 
Together, these rewards encourage models to produce responses that are not only individually appropriate but also coherent at the sequence level. 
To support the two training stages, we construct \textbf{StreamPro-SFT-63K} and \textbf{StreamPro-RL-3K}.
Besides, we further introduce \textbf{StreamPro-Bench}, which organizes proactive capabilities into three complementary dimensions: \emph{Perceptual Understanding}, \emph{Temporal Reasoning}, and \emph{Anticipatory Assistance}. 
The benchmark covers 7 representative tasks across 577 videos and 1,285 question-answer pairs. 
To evaluate proactive response behavior, we propose \textit{StreamPro-F1}, which jointly considers semantic correctness, response timing, and response frequency, and uses task-specific temporal windows to reflect different timing requirements across tasks.
Experiments demonstrate that StreamPro substantially improves proactive streaming video understanding while preserving performance on related tasks. 
On StreamPro-Bench, StreamPro achieves an average StreamPro-F1 of 32.8, compared with 9.2 for the strongest baseline. 
It also maintains strong performance on real-time streaming benchmarks, achieving 79.8 on StreamingBench-RTVU, as well as competitive performance on offline video understanding tasks.

Our contributions are summarized as follows:
(1) We propose \textbf{StreamPro}, a two-stage training framework that addresses key challenges in proactive learning. It combines \textit{CB-Stream Loss} for mitigating supervision imbalance during SFT with GRPO-based training using a refined \textit{RM reward} and a rubric-based \textit{SQ reward} for response-level and sequence-level optimization. We also construct the \textit{StreamPro-SFT-63K} and \textit{StreamPro-RL-3K} datasets.
(2) We introduce \textbf{StreamPro-Bench}, a multi-dimensional benchmark for proactive streaming video understanding, covering \emph{Perceptual Understanding}, \emph{Temporal Reasoning}, and \emph{Anticipatory Assistance} across 7 representative tasks, together with \textbf{StreamPro-F1}, a timing-aware metric that accounts for semantic correctness, response timing, and response frequency.
(3) Experiments demonstrate the effectiveness of StreamPro. It achieves an average StreamPro-F1 of 32.8 on StreamPro-Bench, substantially outperforming existing baselines while maintaining strong real-time streaming video understanding performance.

\begin{table*}[t]
    \centering
    \caption{
        Comparison of proactive benchmarks. Len. denotes average video length, and Proact. Ratio denotes the percentage of proactive questions. PU: Perceptual Understanding, TR: Temporal Reasoning, and AA: Anticipatory Assistance. $\triangle$ indicates partial coverage.
    }
    \label{tab:comparison_bench}
    \begin{adjustbox}{max width=\linewidth}
    \small
    
    \begin{tabular}{lccccccc}
    \toprule
    \textbf{Benchmark} & \textbf{\#Videos} & \textbf{\#Samples} &
    \textbf{Len.(s)} & \textbf{Proact. Ratio(\%)} &
    \textbf{PU} & \textbf{TR} & \textbf{AA} \\
    \midrule
    StreamingBench~\citep{streamingbench}
        & 900 & 4,500 & 243.1 & 5.5 & \cmark & $\triangle$ & \xmark \\
    OVO-Bench~\citep{ovobench}
        & 644 & 2,814 & 428.9 & 10.5 & \cmark & $\triangle$ & \xmark \\
    ProactiveVideoQA~\citep{proactivevideoqa}
        & 1,377 & 1,427 & 124.8 & 100.0 & \cmark & \xmark & \xmark \\
    Omni-MMI~\citep{ominimmi}
        & 1,121 & 2,290 & 324.3 & 22.0 & \cmark & $\triangle$ & \xmark \\
    OmniStar~\citep{livestar}
        & 1,000 & 1,000 & -- & 60.0
        & \cmark & $\triangle$ & \xmark \\
    OmniPro~\citep{omnipro}
        & 1,262 & 2,700 & 189.0 & 100.0 & \cmark & \xmark & $\triangle$ \\
    OmniInteract~\citep{omniinteract}
        & 250 & 1,102 & -- & 31.2
        & \cmark & \xmark & $\triangle$ \\
    \rowcolor{our_color!10}
    \textbf{StreamPro-Bench}
        & 577 & 1,285 & 130.0 & 100.0 & \cmark & \cmark & \cmark \\
    \bottomrule
    \end{tabular}
    \end{adjustbox}
\end{table*}

\section{Related Work}
\label{sec:related_work}

\noindent\textbf{Proactive Streaming Video Understanding.}
Existing approaches to proactive streaming video understanding can be broadly categorized into two groups. 
(1)~\textbf{\emph{Module-based Proactive Models}} introduce additional modules to explicitly control response timing and activation, often relying on heuristics or task-specific components~\citep{querystream,streammind,thinking-qwenvl,vispeak}.
For example, Dispider~\citep{dispider} decomposes streaming interaction into perception, decision, and reaction modules.
StreamBridge~\citep{streambridge} augments offline VideoLLMs with a lightweight external gating model.
StreamAgent~\citep{streamagent} constructs a complete agent framework for temporal decision-making.
(2)~\textbf{\emph{Token-based End-to-End Proactive Model}} use spatial tokens to integrate proactive capability directly into a single model.
A key challenge in this paradigm is the imbalance between \textit{silence} and \textit{response} supervision signals, which biases the model toward remaining silent excessively.
VideoLLM-Online~\citep{videollmonline} mitigates the imbalance by controlling the activation threshold of the EOS token during inference.
Streamo~\citep{streamo} further alleviates this issue.
To better address this issue, we propose CB-Stream Loss, a simple yet effective loss that re-weights streaming control tokens based on their effective frequency.

Recently, several works have explored RL to optimize proactive behavior in streaming settings. Existing rewards emphasize response timing and correctness~\citep{thinkstream} and repetition control~\citep{mmduet2}, but provide limited supervision for overall sequence organization. Strict response matching in TOM~\citep{tom} also gives no credit for partial correctness. We combine a refined RM reward for denser feedback with a rubric-based SQ reward for overall sequence quality.

\noindent\textbf{Proactive Streaming Video Benchmark.}
With the rapid progress of streaming video understanding, many benchmarks have been proposed to evaluate proactive capabilities.
StreamingBench evaluates proactive models through alert-based questions, where models are required to respond upon observing visual evidence in the video~\citep{streamingbench}.
OVO-Bench~\citep{ovobench} evaluates proactive models through three tasks: repetition event counting, sequential step recognition, and clue-revealing response.
ProactiveVideoQA~\citep{proactivevideoqa} is the first benchmark specifically designed for proactive tasks, and it introduces PAUC evaluation metric.
Omni-MMI~\citep{ominimmi} aims for comprehensive multimodal interaction evaluation by introducing audio inputs and multi-turn dialogue tasks.
Compared with existing benchmarks, StreamPro-Bench covers a broader range of tasks that require models to reason about temporal relationships, identify appropriate assistance opportunities from evolving visual context, and guide users through subsequent actions. 
\section{StreamPro-Bench}
\label{sec:benchmark}
In this section, we introduce \textbf{StreamPro-Bench}. 
We first define the task taxonomy across three dimensions (Section~\ref{subsec:task_taxonomy}). We then detail the data construction pipeline and benchmark statistics (Section~\ref{subsec:benchmark_construction}). Finally, we present the \textbf{StreamPro-F1 Score}, a tailored evaluation metric that jointly assesses temporal alignment and semantic correctness (Section~\ref{subsec:eval_metric}).

\begin{figure*}[t]
  \centering
  \includegraphics[width=0.9\linewidth]{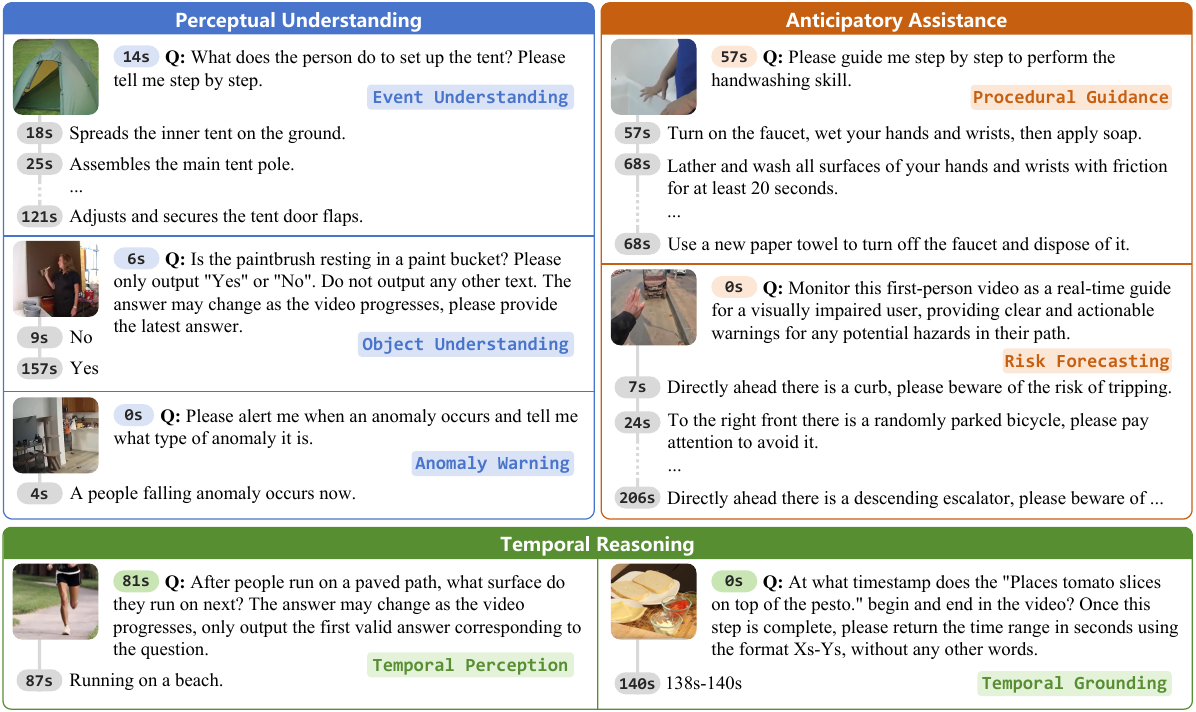}
  \caption{
    Task case illustration of StreamPro-Bench. 
    It contains 7 tasks categorized into three major dimensions: \textcolor[HTML]{4874cb}{Perceptual Understanding}, \textcolor[HTML]{588e32}{Temporal Reasoning}, and \textcolor[HTML]{c65f10}{Anticipatory Assistance}. 
    The timestamps of the user query and the ground truth are explicitly annotated along the timelines. 
  }
  \label{fig:benchmark_task_cases}
\end{figure*}

\subsection{Task Taxonomy}
\label{subsec:task_taxonomy}

We argue that a strong proactive model should exhibit capabilities across three dimensions: Perceptual Understanding, Temporal Reasoning, and Anticipatory Assistance. 
Across these three dimensions, we define 7 specific tasks, as illustrated in Figure~\ref{fig:benchmark_task_cases}.

\noindent\textbf{Perceptual Understanding.} 
This dimension evaluates the foundational ability to continuously perceive the states, dynamics, and changes of entities within streaming video inputs. 
We assess this capability through three tasks: (1) \textit{Event Understanding} (EU): describe the continuous steps of an evolving event as it unfolds (e.g., the sequential actions of a person setting up a tent). 
(2) \textit{Object Understanding} (OU): track the status of specific objects, requiring the model to instantly confirm when a state transition occurs. 
(3) \textit{Anomaly Warning} (AW): trigger immediate warnings upon detecting sudden anomalous events, such as a person falling or a fire breaking out.

\noindent\textbf{Temporal Reasoning.} 
Building on foundational perception, this dimension assesses whether models can track when events happen and how they depend on each other in time. 
It includes two tasks: (1) \textit{Temporal Perception} (TP): given a target event, identify which events occur after it, testing fine-grained chronological understanding.
(2) \textit{Temporal Grounding} (TG): output the start and end timestamps of the given event in a streaming setting as soon as the event concludes.

\noindent\textbf{Anticipatory Assistance.} 
This dimension evaluates whether models can anticipate upcoming task-relevant events from the evolving visual context and provide timely guidance before the next action or potential risk unfolds.
It comprises two tasks: (1) \textit{Procedural Guidance} (PG): given a goal, identify which procedural step the user is currently performing, track its progress, infer the next step, and promptly provide appropriate guidance once the step is completed.
(2) \textit{Risk Forecasting} (RF): in first-person navigation scenarios from the perspective of a visually impaired person, anticipate potential accidents roughly 3 seconds before they occur, warn the user, and provide actionable navigation advice.

\subsection{Benchmark Construction}
\label{subsec:benchmark_construction}

\begin{figure}[t]
  \centering
  \includegraphics[width=0.8\linewidth]{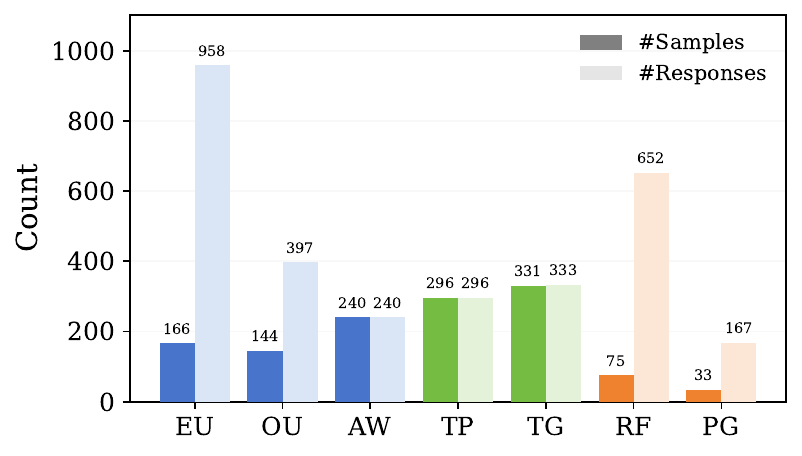}
  \caption{Number of samples and responses of each StreamPro-Bench task.}
  \label{fig:bench_statistics}
\end{figure}

We construct data for all 7 tasks using a pipeline that includes a two-agent verification loop and human refinement.
Given the complexity of Risk Forecasting task, every sample is manually reviewed, and incorrect annotations are manually redone.
Further details are provided in Supplementary Material.
In total, StreamPro-Bench comprises 577 videos and 1,285 QA pairs. 
Figure~\ref{fig:bench_statistics} shows the number of instances for each task and the total number of corresponding responses, as a single question may be associated with multiple answers.
Although Anticipatory Assistance contains only 108 queries, each query corresponds to an average of 7.6 ground-truth response instances, resulting in 819 response instances in total.
The comparison with current mainstream benchmarks is shown in Table~\ref{tab:comparison_bench}.

\subsection{Evaluation Protocol}
\label{subsec:eval_metric}

\noindent\textbf{Evaluation Metric.}
We introduce \textbf{StreamPro-F1}, which jointly evaluates response timing and semantic correctness while penalizing excessive and missed responses.
For a predicted response and a ground-truth response, the \textbf{semantic score} $S_{\mathrm{sem}}$ is evaluated by an LLM judge for most tasks. For Temporal Grounding, we use Intersection over Union (IoU).
The \textbf{timing score} $S_{\mathrm{time}}$ is evaluated using the temporal distance between the predicted response time $t_{\mathrm{pred}}$ and the optimal response timing $t_{\mathrm{gt}}$. It decreases linearly to zero as this distance reaches the task-specific tolerance $\tau$. The joint score $S$ is the product of the timing and semantic scores, so a response receives a high score only when it is both timely and semantically accurate:
\begin{equation}
\begin{aligned}
S_{\mathrm{time}} &=
\max\left(0,1-\frac{\operatorname{dist}(t_{\mathrm{pred}},t_{\mathrm{gt}})}{\tau}\right),\\
S &= S_{\mathrm{time}}\cdot S_{\mathrm{sem}}.
\end{aligned}
\end{equation}
At the trajectory level, each matched pair contributes its joint score $S_i$. We compute score-weighted precision $P$ and recall $R$ using the total numbers of predicted and ground-truth responses, denoted by $N_{\mathrm{pred}}$ and $N_{\mathrm{gt}}$. Excessive responses lower precision, while missed ground-truth responses lower recall. The final StreamPro-F1 score is the harmonic mean of $P$ and $R$:
\begin{equation}
P=\frac{\sum_i S_i}{N_{\mathrm{pred}}},
\quad
R=\frac{\sum_i S_i}{N_{\mathrm{gt}}},
\quad
F_1=\frac{2PR}{P+R}.
\end{equation}
\noindent StreamPro-F1 aligns well with human preferences. See detailed information in Supplementary Material.

\noindent{\textbf{Evaluation Method.}}
For proactive models, we evaluate only those with officially released proactive inference scripts to ensure fair and reproducible evaluation~\citep{videollmonline,mmduet,mmduet2,minicpm_o_45,streamo}.
For offline models~\citep{qwen25vl,qwen3vl}, we design a separate decision-tree-based frame-by-frame evaluation protocol to assess their proactive capabilities.
Further details regarding the evaluation protocol are provided in Supplementary Material.

\section{StreamPro Training Framework}
\label{sec:method}
\begin{figure*}[t]
  \centering
  \includegraphics[width=0.9\linewidth]{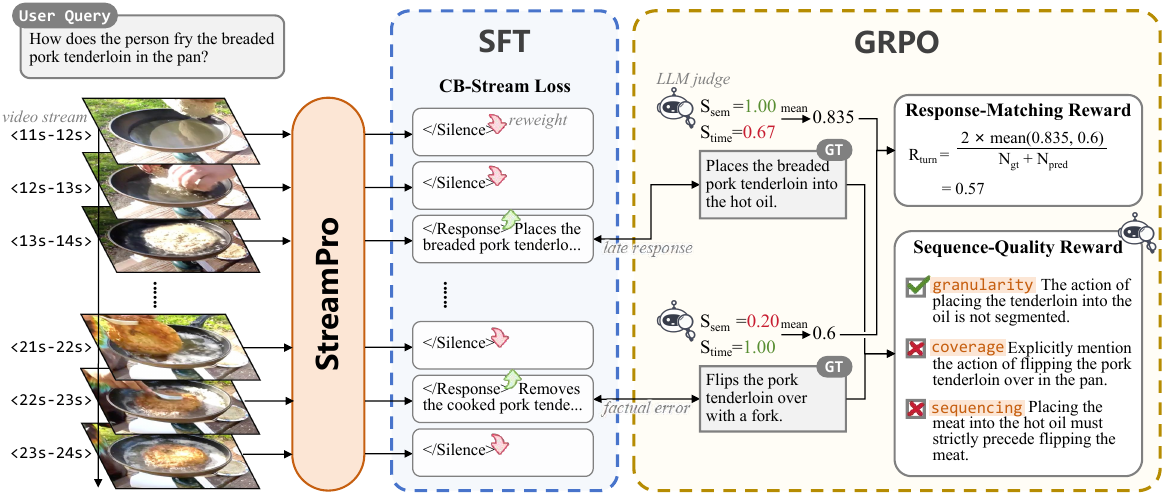}
  \caption{
    Overview of the StreamPro training framework. 
    \textbf{In the SFT stage}, the CB-Stream Loss adjusts the weights of the \texttt{</Silence>} and \texttt{</Response>} decision tokens according to their effective sample sizes in each batch.
    \textbf{In the GRPO stage}, the RM reward combines the semantic correctness assessed by an LLM judge with the timeliness decayed based on temporal distance. Here, $N_{\mathrm{gt}}$ and $N_{\mathrm{pred}}$ denote the respective total numbers of ground truth and predicted events, which are both 2 in this illustrated example. The rubric-based SQ reward assesses granularity, sequencing, and coverage across the complete response sequence.
  }
  \label{fig:method}
\end{figure*}

This section introduces our two-stage StreamPro training framework (Figure~\ref{fig:method}): Section~\ref{subsec:sft} details the SFT stage, Section~\ref{subsec:rl} the RL stage, and Section~\ref{subsec:training_data} the training data for both stages.

\subsection{Supervised Fine-Tuning}
\label{subsec:sft}

\noindent\textbf{Proactive Streaming Decision Format.}
At each time step, the model either remains silent or generates a response conditioned on the current video context.
Specifically, the model outputs \texttt{</Silence>} when no response is required, and \texttt{</Response>} followed by the corresponding textual answer when a response is triggered. We denote the decision token set as $\mathcal{S} = \{\texttt{</Silence>}, \texttt{</Response>}\}$, and the standard language token set as $\mathcal{T}$.

\noindent\textbf{CB-Stream Loss.}
Proactive streaming video data exhibits a severe imbalance between response and silence signals, which causes standard cross-entropy training to favor conservative policies that over-predict silence.
To mitigate this issue, we adopt a simple yet effective class-balanced reweighting strategy~\citep{cb} based on the effective number of samples.
For each decision token class $k \in \mathcal{S}$, we define the effective sample size $E_k$ and corresponding class-balanced weight $\hat{w}_k^{\mathrm{CB}}$ as:
\begin{equation}
  E_k = \frac{1 - \beta^{n_k}}{1 - \beta}, 
  \quad
  \hat{w}_k^{\mathrm{CB}} = \frac{1/E_k}{\sum_{j \in \mathcal{S}} 1/E_j} \cdot |\mathcal{S}|,
\end{equation}
where $n_k$ is the frequency of class $k$ computed over the current batch, and $\beta \in [0,1)$ controls the degree of reweighting.
We further introduce a constant scaling factor $\lambda_{\mathrm{text}}$ to balance optimization between decision tokens and language tokens. The final training objective is defined as:
\begin{equation}
  \mathcal{L}_{\mathrm{CB}} = \frac{1}{N} \sum_{i=1}^{N}
  w_i^{\mathrm{CB}} \cdot \bigl[-\log p_i\bigr], \quad
  w_i^{\mathrm{CB}} =
  \begin{cases}
    \hat{w}_{y_i}^{\mathrm{CB}}, & y_i \in \mathcal{S}, \\
    \lambda_{\mathrm{text}}, & y_i \in \mathcal{T}.
  \end{cases}
\end{equation}

\subsection{Reinforcement Learning}
\label{subsec:rl}

Although SFT equips the model with basic streaming and proactive response capabilities, it is limited by exposure bias: the model is trained on ground-truth decisions but must rely on its own decisions at inference, so errors accumulate over the video stream.
To mitigate this, we employ GRPO~\citep{grpo} with three rewards: a format reward, a refined \emph{Response-Matching (RM)} reward, and a rubric-based \emph{Sequence-Quality (SQ)} reward.
The total reward is:
\begin{equation}
  R = w_{\mathrm{fmt}} R_{\mathrm{fmt}} + w_{\mathrm{RM}} R_{\mathrm{RM}} + w_{\mathrm{SQ}} R_{\mathrm{SQ}},
\end{equation}
where the $w$ terms are the corresponding reward weights.

\noindent\textbf{Format Reward.}
The format reward checks the output at each time step: a standalone \texttt{</Silence>} token receives a score of $1$, as does a \texttt{</Response>} token followed by non-empty text; all other outputs receive $0$. We average the step scores to obtain $R_{\mathrm{fmt}}$.

\noindent\textbf{RM Reward.}
Although StreamPro-F1 aligns well with human preferences, it is too strict to serve directly as an RL reward: its multiplicative score and narrow temporal tolerance yield sparse rewards, giving the model little guidance for improving partially correct responses. We therefore make two changes. (1)~\emph{Denser scoring}: we use the mean of the timing and semantic scores instead of their product, so a weak component no longer erases the other. (2)~\emph{Wider tolerance}: we increase the temporal tolerance $\tau$ to widen the matching window and slow the decay of the timing score, allowing small temporal deviations to receive partial credit. We use a single $\tau$ across all tasks to keep the reward design simple and task-agnostic. 

The reward retains the F1 penalty for extra and missed responses:
\begin{equation}
  \label{eq:rm_reward}
  R_{\mathrm{RM}} = \frac{2 \sum_i S'_i}{N_{\mathrm{pred}} + N_{\mathrm{gt}}}, \quad S'_i = \frac{S_{\mathrm{time}, i} + S_{\mathrm{sem}, i}}{2}.
\end{equation}

\noindent\textbf{SQ Reward.}
The RM reward scores individual responses but cannot assess whether they form a coherent sequence.
In Event Understanding and Procedural Guidance, the model must choose an appropriate level of detail, cover the key points, and preserve their order. We use the SQ reward to assess these properties over the complete sequence. The rubric covers three properties: (1)~\textbf{Granularity}, responses are neither too fragmented nor too coarse relative to event duration; (2)~\textbf{Sequencing}, chronological consistency; (3)~\textbf{Coverage}, essential points are included while hallucinations are penalized. Concretely, before training, an LLM designer generates a rubric of $N_c$ binary checkpoints from each sample's question and ground truth; during GRPO, an LLM evaluator checks the generated sequence against each checkpoint and assigns $c_i \in \{0, 1\}$. The final reward averages the checkpoint scores:
\begin{equation}
  R_{\mathrm{SQ}} = \frac{1}{N_c} \sum_{i=1}^{N_c} c_i.
\end{equation}

\subsection{Training Data}
\label{subsec:training_data}

We construct two datasets using the StreamPro-Bench data pipeline: \textbf{StreamPro-SFT-63K} and \textbf{StreamPro-RL-3K}. 
We enforce strict data isolation throughout dataset construction to ensure that all training data are fully disjoint from the benchmark evaluation set, preventing any potential data leakage.
We then use these datasets in the StreamPro training framework. 
In the SFT stage, we jointly train on real-time streaming and proactive data, including TimeChat-Online-139K~\cite{timechat-online}, VideoChat-Flash-3K~\cite{videochat-flash}, StreamPro-SFT-63K, and 287K filtered samples from Streamo-Instruct-465K~\cite{streamo}. 
In the RL stage, we focus on proactive tasks and train the model using StreamPro-RL-3K.
The training set composition are described in Supplementary Material.

\section{Experiments}
\label{sec:experiments}

\begin{table*}[t]
    \centering
    \small
    \setlength{\tabcolsep}{3pt}
    \caption{Performance on proactive tasks across StreamPro-Bench and OVO-Bench. PU: Perceptual Understanding, TR: Temporal Reasoning, AA: Anticipatory Assistance.}
    \label{tab:proactive_result_streampro_ovo}
    \adjustbox{max width=\linewidth}{
    \begin{tabular}{l|c|*{10}{c}|*{3}{c}}
        \toprule
        \multirow{3}{*}{\textbf{Methods}} & \multirow{3}{*}{\textbf{Params}}
        & \multicolumn{10}{c}{\textbf{StreamPro-Bench}}
        & \multicolumn{3}{|c}{\textbf{OVO-Bench}} \\
        \cmidrule(lr){3-12} \cmidrule(lr){13-15}

        & & \multicolumn{3}{c}{\textbf{PU}}
        & \multicolumn{3}{c}{\textbf{TR}}
        & \multicolumn{3}{c}{\textbf{AA}}
        & \multirow{2}{*}{\textbf{Avg. F1}}
        & \multicolumn{3}{c}{\textbf{FAR}} \\
        \cmidrule(lr){3-5} \cmidrule(lr){6-8}
        \cmidrule(lr){9-11} \cmidrule(lr){13-15}

        & & Time & Sem. & F1
        & Time & Sem. & F1
        & Time & Sem. & F1
        & & Time & Sem. & F1 \\

        \midrule
        \multicolumn{15}{c}{\textit{Offline Models + Proactive Prompt}} \\
        \midrule
        Qwen2.5-VL-7B~\citep{qwen25vl} & 7B & 8.7 & 16.3 & 4.2 & 5.0 & 0.9 & 0.6 & 7.5 & 6.1 & 4.4 & 3.1 & - & - & - \\
        Qwen3-VL-8B~\citep{qwen3vl} & 8B & 59.5 & 50.1 & 5.7 & 28.8 & 17.6 & 0.9 & 33.0 & 17.4 & 3.1 & 3.2 & - & - & - \\

        \midrule
        \multicolumn{15}{c}{\textit{Open-Source Proactive Models}} \\
        \midrule
        VideoLLM-Online~\citep{videollmonline} & 8B & 14.6 & 1.3 & 0.3 & 0.6 & 0.1 & 0.0 & 21.5 & 8.0 & \textbf{8.8} & 3.0 & 4.4 & 0.4 & 0.1 \\
        MMDuet~\citep{mmduet} & 7B & 50.5 & 34.7 & 9.1 & 72.9 & 21.4 & 1.6 & 33.6 & 15.8 & 3.8 & 4.8 & 42.5 & 19.2 & 5.7 \\
        MMDuet2~\citep{mmduet2} & 3B & 41.0 & 29.2 & 9.7 & 49.3 & 31.8 & 3.4 & 27.3 & 9.7 & 3.8 & 5.6 & 33.5 & 18.1 & 6.5 \\
        MiniCPM-o-4.5~\citep{minicpm_o_45} & 9B & 18.8 & 16.9 & 13.2 & 2.4 & 0.4 & 0.1 & 7.5 & 6.1 & 4.4 & 5.9 & 15.6 & 10.6 & 8.4 \\
        Streamo~\citep{streamo} & 3B & 24.6 & 20.5 & 9.0 & 27.9 & 19.5 & 14.3 & 9.2 & 4.0 & 4.4 & 9.2 & 11.5 & 7.7 & 5.4 \\
        \rowcolor{our_color!10}
        StreamPro-SFT & 3B & 51.8 & 44.1 & 13.7 & 45.8 & 26.8 & 22.0 & 42.0 & 20.3 & 4.3 & 13.3 & 41.6 & 25.4 & 9.5 \\
        \rowcolor{our_color!10}
        StreamPro-GRPO & 3B & 55.6 & 51.3 & \underline{35.0} & 58.7 & 47.4 & \underline{34.7} & 14.2 & 8.9 & 5.6 & \underline{25.1} & 37.5 & 20.8 & 14.4 \\
        \rowcolor{our_color!20}
        StreamPro-SFT & 4B & 67.2 & 61.8 & 23.8 & 46.3 & 33.5 & 29.1 & 43.6 & 21.9 & 6.5 & 19.8 & 41.6 & 32.2 & \underline{17.3} \\
        \rowcolor{our_color!20}
        StreamPro-GRPO & 4B & 65.7 & 60.6 & \textbf{44.8} & 68.3 & 61.0 & \textbf{45.1} & 20.8 & 15.6 & \underline{8.5} & \textbf{32.8} & 44.2 & 33.9 & \textbf{20.6} \\
        \bottomrule
    \end{tabular}
    }
\end{table*}

\subsection{Experiment Settings}

\noindent\textbf{Benchmarks.}
We evaluate our proposed model across three distinct categories of tasks:
    \textbf{(1) \emph{Proactive Tasks.}} We evaluate on StreamPro-Bench (SPB) and Forward Active Responding (FAR) tasks from OVO-Bench~\cite{ovobench}. Performance is measured using StreamPro-F1. We compare our approach with baseline models that provide open-source scripts for proactive inference~\cite{videollmonline,mmduet,mmduet2,minicpm_o_45,streamo}.
    \textbf{(2) \emph{Real-time Streaming Tasks.}} We evaluate on Real-Time Visual Perception (RTVP) from OVO-Bench, alongside Real-Time Visual Understanding (RTVU) and Contextual Understanding (CU) from StreamingBench~\cite{streamingbench}. We exclude the PO task from RTVU as it evaluates proactive behavior, and compute the RTVU average over the remaining tasks.
    For these tasks, we compare against 11 representative streaming models~\cite{flash-vstream,timechat-online,streamforest,streamingvlm,dispider,vispeak,streamagent,querystream,thinking-qwenvl,videollmonline,streamo}.
    \textbf{(3) \emph{Offline Tasks.}} We evaluate on VideoMME~\cite{videomme} and LongVideoBench~\cite{longvideobench}.

\noindent\textbf{Implementation Details.}
We employ Qwen2.5-VL-3B~\cite{qwen25vl} and Qwen3-VL-4B~\cite{qwen3vl} as our backbone models, and ablate on the 3B model.
During SFT, we train both the projector and the LLM for 1 epoch using 64 H100 GPUs for 24 hours, with a learning rate of $1\times10^{-5}$ and a batch size of 512. 
The reweighting hyperparameter $\beta$ is set to 0.9999, and the text scaling factor $\lambda_{\text{text}}$ is set to 2.
During RL, we implement the GRPO pipeline using veRL~\cite{verl} and vLLM~\cite{vllm}, training for 1 epoch using 8 H100 GPUs for 28 hours with a learning rate of $10^{-6}$ and a global batch size of 16. 
We sample $G=8$ generations per video context with a temperature of 1.0. 
For the GRPO rewards, we set the temporal tolerance $\tau = 8$, along with the reward weights $w_{\mathrm{fmt}} = 0.1$, $w_{\mathrm{RM}} = 0.45$, and $w_{\mathrm{SQ}} = 0.45$.
All videos are sampled at 1 FPS. 
During inference, we employ a sliding window of 200 dialogue turns to improve inference efficiency.

\subsection{Main Results}

\begin{table*}[t]
    \centering
    \small
    \caption{Performance on real-time streaming tasks and offline video benchmarks. 
    \textsuperscript{$\dagger$} denotes results from our evaluation of the official Streamo-3B checkpoint.}
    \label{tab:realtime_offline_results}
    \adjustbox{max width=\linewidth}{
    \begin{tabular}{l|c|c|cc|c|c}
        \toprule
        \multirow{2}{*}{\textbf{Methods}} & \multirow{2}{*}{\textbf{Params}}
        & \textbf{OVO-Bench}
        & \multicolumn{2}{c|}{\textbf{StreamingBench}}
        & \multirow{2}{*}{\textbf{VideoMME}}
        & \multirow{2}{*}{\textbf{LongVideoBench}} \\
        \cmidrule(lr){3-3} \cmidrule(lr){4-5}
        & & \textbf{RTVP} & \textbf{RTVU} & \textbf{CU}
        & & \\
        \midrule
        \multicolumn{7}{c}{\textit{Non-Proactive Streaming Models}} \\
        \midrule
        Flash-VStream~\citep{flash-vstream} & 7B & 28.4 & 25.9 & 25.6 & - & - \\
        TimeChat-Online~\citep{timechat-online} & 7B & 61.9 & 75.3 & 37.5 & 62.4 & 55.4 \\
        StreamForest~\citep{streamforest} & 7B & 61.2 & 77.3 & - & 61.4 & - \\
        StreamingVLM~\citep{streamingvlm} & 7B & 62.0 & - & - & 65.1 & 59.0 \\

        \midrule
        \multicolumn{7}{c}{\textit{Module-based Proactive Models}} \\
        \midrule
        Dispider~\citep{dispider} & 7B & 54.5 & 67.5 & 34.0 & 57.2 & - \\
        ViSpeak~\citep{vispeak} & 7B & 66.3 & 74.5 & 39.9 & 58.0 & - \\
        StreamAgent~\citep{streamagent} & 7B & 61.3 & 74.6 & 36.5 & 62.9 & 57.9 \\
        QueryStream~\citep{querystream} & 7B & 63.1 & 75.3 & - & 63.8 & 57.3 \\
        Thinking-QwenVL~\citep{thinking-qwenvl} & 7B & 64.7 & 71.6 & - & \underline{67.7} & \textbf{62.0} \\

        \midrule
        \multicolumn{7}{c}{\textit{Token-based End-to-End Proactive Models}} \\
        \midrule
        VideoLLM-Online~\citep{videollmonline} & 8B & 20.8 & 36.0 & 28.1 & - & - \\
        Streamo-3B\textsuperscript{$\dagger$}~\citep{streamo} & 3B & 61.2 & 75.8 & 41.1 & 58.0 & 51.8 \\
        Streamo-7B~\citep{streamo} & 7B & 66.0 & - & - & \textbf{67.9} & 59.2 \\
        \rowcolor{our_color!10}
        StreamPro-SFT & 3B & 62.1 & 77.1 & 40.0 & 60.7 & 54.6 \\
        \rowcolor{our_color!10}
        StreamPro-GRPO & 3B & 64.2 & 75.9 & 39.6 & 60.9 & 54.6 \\
        \rowcolor{our_color!20}
        StreamPro-SFT & 4B & \underline{68.2} & \underline{79.3} & \textbf{46.9} & 66.5 & \underline{61.9} \\
        \rowcolor{our_color!20}
        StreamPro-GRPO & 4B & \textbf{69.7} & \textbf{79.8} & \underline{46.7} & 67.3 & 60.4 \\
        \bottomrule
    \end{tabular}
    }
\end{table*}

\noindent\textbf{Proactive Tasks.} As shown in Table~\ref{tab:proactive_result_streampro_ovo}, both training stages improve proactive performance. On StreamPro-Bench, the 4B SFT and GRPO models achieve F1 scores of 19.8 and 32.8, respectively, substantially exceeding the best previous proactive model's score of 9.2. On OVO-Bench FAR, evaluated with the StreamPro-F1 metric, the 4B model improves from 17.3 after SFT to 20.6 after GRPO, further demonstrating that our performance gains generalize beyond our StreamPro-Bench. 

\noindent\textbf{Real-Time Streaming and Offline Tasks.}
As shown in Table~\ref{tab:realtime_offline_results}, StreamPro-GRPO-4B outperforms the published 7B baselines on real-time streaming tasks (OVO-Bench RTVP and StreamingBench RTVU). Furthermore, it achieves performance comparable to many 7B models on offline benchmarks (VideoMME and LongVideoBench), despite the absence of offline data throughout both training stages. When compared against StreamPro-SFT-4B, GRPO yields slightly reduced scores on CU as well as LongVideoBench. These modest performance drops are reasonable, considering that the RL stage is trained solely using proactive streaming data. Taken together, these results suggest that StreamPro‑GRPO‑4B maintains favorable generalization capabilities beyond proactive tasks.
Detailed benchmark results are provided in Supplementary Material.

\subsection{Ablation Study}

\noindent\textbf{Effect of CB-Stream Loss.}
As shown in Table~\ref{tab:ablation_loss}, replacing the standard CE loss with CB-Stream substantially improves proactive performance, increasing the SPB score from 4.8 to 13.7, while largely preserving performance on OVO-RTVP and VideoMME. For the text-token weight $\lambda_{\text{text}}$, the results with $\lambda_{\text{text}}=1$ and $2$ are comparable overall. Considering the balance between proactive performance and general video understanding, we set $\lambda_{\text{text}}=2$ in our final configuration.

\begin{table}[t]
    \centering
    \small
    \caption{Ablation of the loss function and text token weight $\lambda_{\mathrm{text}}$ during SFT.}
    \label{tab:ablation_loss}
    \begin{adjustbox}{max width=\linewidth}
    \renewcommand{\arraystretch}{1.05}
    \begin{tabular}{l ccc}
        \toprule
        \textbf{Loss} & \textbf{SPB} & \textbf{OVO-RTVP} & \textbf{VideoMME} \\
        \midrule
        CE & 4.8 & \textbf{62.3} & 60.4 \\
        CB-Stream {\scriptsize($\lambda=1$)} & \textbf{13.7} & 61.8 & 60.4 \\
        \rowcolor{our_color!10}
        CB-Stream {\scriptsize($\lambda=2$)} & \underline{13.3} & \underline{62.1} & \underline{60.7} \\
        CB-Stream {\scriptsize($\lambda=3$)} & 12.7 & 60.8 & \textbf{60.9} \\
        \bottomrule
    \end{tabular}
    \end{adjustbox}
\end{table}

\noindent\textbf{Effectiveness of StreamPro-SFT-63K in SFT.}
To verify the quality of the data generated by our pipeline, we conduct an ablation study as shown in Table~\ref{tab:data_ablation}. 
When training only with the 429K open-source data, the model achieves 8.6 on SPB and 60.7 on OVO-RTVP. 
After incorporating the additional StreamPro-SFT-63K data, performance improves substantially, reaching 13.3 on SPB and 62.1 on OVO-RTVP, while also improving VideoMME from 58.5 to 60.7. 
These results demonstrate the effectiveness and high quality of our constructed SFT data in enhancing both proactive and real time streaming tasks.

\begin{table}[t]
    \centering
    \small
    \caption{Ablation study on the effectiveness of StreamPro-SFT-63K data during SFT training.}
    \label{tab:data_ablation}
    \begin{adjustbox}{max width=\linewidth}
    \renewcommand{\arraystretch}{1.05}
    \begin{tabular}{l ccc}
        \toprule
        \textbf{Data} & \textbf{SPB} & \textbf{OVO-RTVP} & \textbf{VideoMME} \\
        \midrule
        Open source 429k & 8.6 & 60.7 & 58.5 \\
        \rowcolor{our_color!10}
        429k+StreamPro-SFT-63k & \textbf{13.3} & \textbf{62.1} & \textbf{60.7} \\
        \bottomrule
    \end{tabular}
    \end{adjustbox}
\end{table}

\noindent\textbf{RM Reward Ablation.}
Compared to StreamPro-F1, the RM reward uses a larger shared tolerance $\tau$ and takes the mean instead of the product of timing and semantic scores to provide denser feedback. As shown in rows 1--3 of Table~\ref{tab:ablation_rl}, increasing $\tau$ progressively improves StreamPro-Bench F1. A larger tolerance widens the matching window and allows responses close to the reference time to receive partial credit. Comparing rows 3 and 5, replacing the product with the mean improves StreamPro-Bench F1 by 3.3 points and OVO-RTVP by 3.6 points, suggesting that averaging provides more effective feedback when either timing or semantic quality is weak.

\noindent\textbf{SQ Reward Ablation.}
The SQ reward evaluates granularity, sequencing, and coverage across the complete response sequence. Comparing rows 3 and 4 of Table~\ref{tab:ablation_rl}, adding SQ improves StreamPro-Bench F1 by 4.2 points and OVO-RTVP by 3.3 points. These gains indicate that proactive training benefits from considering how responses relate to one another, beyond merely scoring each matched response independently.

\begin{table}[t]
    \centering
    \small
    \caption{Ablation of RM score combination, temporal tolerance $\tau$, and the SQ reward during RL.}
    \label{tab:ablation_rl}
    \begin{adjustbox}{max width=\linewidth}
    \begin{tabular}{ccccccc}
        \toprule
        \textbf{No.} & \textbf{Comb.} & $\boldsymbol{\tau}$ & \textbf{SQ Reward} & \textbf{SPB} & \textbf{OVO-RTVP} & \textbf{VideoMME} \\
        \midrule
        1 & Mean & 3 & \cmark & 18.7 & \underline{61.6} & \underline{60.9} \\
        2 & Mean & 5 & \cmark & \underline{22.2} & 61.4 & 60.8 \\
        \rowcolor{our_color!10}
        3 & Mean & 8 & \cmark & \textbf{25.1} & \textbf{64.2} & \underline{60.9} \\
        4 & Mean & 8 & \xmark & 20.9 & 60.9 & 60.6 \\
        5 & Prod. & 8 & \cmark & 21.8 & 60.6 & \textbf{61.0} \\
        \bottomrule
    \end{tabular}
    \end{adjustbox}
\end{table}

\noindent\textbf{Sensitivity to Different LLM Judges.}
To assess the robustness of StreamPro-Bench to the choice of LLM judge, we evaluate StreamPro-GRPO-4B using three different judges: Gemini 2.5 Pro, GPT-5.5, and Gemini 3.1 Pro. As shown in Table~\ref{tab:llm_judge_sensitivity}, the resulting average scores are 32.8, 33.4, and 33.2, respectively, with a maximum difference of only 0.6 points. The small variation across judges indicates that the evaluation results are largely insensitive to the choice of judging model, supporting the robustness of StreamPro-Bench under different LLM evaluators.

\begin{table}[t]
\centering
\small
\setlength{\tabcolsep}{12pt}
\caption{Sensitivity of StreamPro-Bench to different LLM judges.}
\label{tab:llm_judge_sensitivity}
\begin{tabular}{lc}
\toprule
\textbf{LLM Judge} & \textbf{Average Score} \\
\midrule
Gemini 2.5 Pro & 32.8 \\
GPT-5.5 & 33.4 \\
Gemini 3.1 Pro & 33.2 \\
\bottomrule
\end{tabular}
\end{table}

\section{Conclusions}

\label{sec:conclusion}
In this paper, we introduce the StreamPro framework to address the silence--response imbalance and reward sparsity: a CB-Stream loss that balances the two signals during SFT, and an RL stage combining a refined RM reward with a rubric-based SQ reward that assesses sequence-level quality.
We further propose StreamPro-Bench, a comprehensive benchmark to evaluate proactive video understanding along three dimensions: Perceptual Understanding, Temporal Reasoning, and Anticipatory Assistance.
StreamPro substantially improves proactive capability while preserving real-time streaming performance and remaining competitive on offline benchmarks.
StreamPro and StreamPro-Bench together provide a systematic foundation for proactive streaming video understanding, and we hope they facilitate future work toward real-world proactive assistant systems.

\section*{LLM Usage Statement}
Generative AI tools were used to assist with several aspects of this research and manuscript preparation.
(1) Benchmark Data Generation. Google Gemini 2.5 Pro was used to assist in generating data for the construction of StreamPro-Bench, supporting the development of benchmark evaluation instances.
(2) Code Development. Claude Code (Anthropic) and DeepSeek were used to assist with research code development, including code generation, implementation refinement, and debugging.
(3) Manuscript Preparation. ChatGPT (OpenAI) was used for language polishing, grammar checking, and improving the clarity and organization of the manuscript, including the Abstract, Introduction, and Conclusion.
{
    \small
    \bibliographystyle{ieeenat_fullname}
    \bibliography{iclr2027_conference}

@article{omniinteract,
  title={OmniInteract: Benchmarking Real-World Streaming Interaction for Real-Time Omnimodal Assistants},
  author={Lu, Xudong and Li, Xueying and Wang, Annan and Bo, Yang and Chen, Jinpeng and Li, Zengliang and Yang, Nianzu and Liu, Rui and Yang, Xue and Hou, Jingwen and others},
  journal={arXiv preprint arXiv:2605.26485},
  year={2026}
}

@article{omnipro,
  title={OmniPro: A Comprehensive Benchmark for Omni-Proactive Streaming Video Understanding},
  author={Zhao, Ruixiang and Yang, Jie and Xin, Zijie and Wang, Tianyi and Rao, Fengyun and LYU, Jing and Li, Xirong},
  journal={arXiv preprint arXiv:2605.18577},
  year={2026}
}

@article{verl,
  title   = {HybridFlow: A Flexible and Efficient RLHF Framework},
  author  = {Guangming Sheng and Chi Zhang and Zilingfeng Ye and Xibin Wu and Wang Zhang and Ru Zhang and Yanghua Peng and Haibin Lin and Chuan Wu},
  year    = {2024},
  journal = {arXiv preprint arXiv: 2409.19256}
}

@inproceedings{vllm,
  title={Efficient Memory Management for Large Language Model Serving with PagedAttention},
  author={Woosuk Kwon and Zhuohan Li and Siyuan Zhuang and Ying Sheng and Lianmin Zheng and Cody Hao Yu and Joseph E. Gonzalez and Hao Zhang and Ion Stoica},
  booktitle={Proceedings of the ACM SIGOPS 29th Symposium on Operating Systems Principles},
  year={2023}
}

@inproceedings{videollmonline,
  title={Videollm-online: Online video large language model for streaming video},
  author={Chen, Joya and Lv, Zhaoyang and Wu, Shiwei and Lin, Kevin Qinghong and Song, Chenan and Gao, Difei and Liu, Jia-Wei and Gao, Ziteng and Mao, Dongxing and Shou, Mike Zheng},
  booktitle={Proceedings of the IEEE/CVF Conference on Computer Vision and Pattern Recognition},
  pages={18407--18418},
  year={2024}
}

@article{mmduet,
  title={Videollm knows when to speak: Enhancing time-sensitive video comprehension with video-text duet interaction format},
  author={Wang, Yueqian and Meng, Xiaojun and Wang, Yuxuan and Liang, Jianxin and Wei, Jiansheng and Zhang, Huishuai and Zhao, Dongyan},
  journal={arXiv preprint arXiv:2411.17991},
  volume={1},
  number={3},
  pages={5},
  year={2024}
}

@article{mmduet2,
  title={MMDuet2: Enhancing Proactive Interaction of Video MLLMs with Multi-Turn Reinforcement Learning},
  author={Wang, Yueqian and Liu, Songxiang and Wang, Disong and Xu, Nuo and Wan, Guanglu and Zhang, Huishuai and Zhao, Dongyan},
  journal={arXiv preprint arXiv:2512.06810},
  year={2025}
}

@inproceedings{vispeak,
  title={Vispeak: Visual instruction feedback in streaming videos},
  author={Fu, Shenghao and Yang, Qize and Li, Yuan-Ming and Peng, Yi-Xing and Lin, Kun-Yu and Wei, Xihan and Hu, Jian-Fang and Xie, Xiaohua and Zheng, Wei-Shi},
  booktitle={Proceedings of the IEEE/CVF International Conference on Computer Vision},
  pages={21778--21788},
  year={2025}
}

@article{streambridge,
  title={Streambridge: Turning your offline video large language model into a proactive streaming assistant},
  author={Wang, Haibo and Feng, Bo and Lai, Zhengfeng and Xu, Mingze and Li, Shiyu and Ge, Weifeng and Dehghan, Afshin and Cao, Meng and Huang, Ping},
  journal={arXiv preprint arXiv:2505.05467},
  year={2025}
}

@article{streamo,
  title={Streaming Video Instruction Tuning},
  author={Xia, Jiaer and Chen, Peixian and Zhang, Mengdan and Sun, Xing and Zhou, Kaiyang},
  journal={arXiv preprint arXiv:2512.21334},
  year={2025}
}

@inproceedings{timechat-online,
  title={Timechat-online: 80\% visual tokens are naturally redundant in streaming videos},
  author={Yao, Linli and Li, Yicheng and Wei, Yuancheng and Li, Lei and Ren, Shuhuai and Liu, Yuanxin and Ouyang, Kun and Wang, Lean and Li, Shicheng and Li, Sida and others},
  booktitle={Proceedings of the 33rd ACM International Conference on Multimedia},
  pages={10807--10816},
  year={2025}
}

@article{flash-vstream,
  title={Flash-vstream: Memory-based real-time understanding for long video streams},
  author={Zhang, Haoji and Wang, Yiqin and Tang, Yansong and Liu, Yong and Feng, Jiashi and Dai, Jifeng and Jin, Xiaojie},
  journal={arXiv preprint arXiv:2406.08085},
  year={2024}
}

@article{streamforest,
  title={Streamforest: Efficient online video understanding with persistent event memory},
  author={Zeng, Xiangyu and Qiu, Kefan and Zhang, Qingyu and Li, Xinhao and Wang, Jing and Li, Jiaxin and Yan, Ziang and Tian, Kun and Tian, Meng and Zhao, Xinhai and others},
  journal={arXiv preprint arXiv:2509.24871},
  year={2025}
}

@article{streamagent,
  title={Streamagent: Towards anticipatory agents for streaming video understanding},
  author={Yang, Haolin and Tang, Feilong and Zhao, Lingxiao and An, Xiang and Hu, Ming and Li, Huifa and Zhuang, Xinlin and Lu, Yifan and Zhang, Xiaofeng and Swikir, Abdalla and others},
  journal={arXiv preprint arXiv:2508.01875},
  year={2025}
}

@article{gemini2.5,
  title={Gemini 2.5: Pushing the frontier with advanced reasoning, multimodality, long context, and next generation agentic capabilities},
  author={Comanici, Gheorghe and Bieber, Eric and Schaekermann, Mike and Pasupat, Ice and Sachdeva, Noveen and Dhillon, Inderjit and Blistein, Marcel and Ram, Ori and Zhang, Dan and Rosen, Evan and others},
  journal={arXiv preprint arXiv:2507.06261},
  year={2025}
}

@article{gpt4,
  title={Gpt-4 technical report},
  author={Achiam, Josh and Adler, Steven and Agarwal, Sandhini and Ahmad, Lama and Akkaya, Ilge and Aleman, Florencia Leoni and Almeida, Diogo and Altenschmidt, Janko and Altman, Sam and Anadkat, Shyamal and others},
  journal={arXiv preprint arXiv:2303.08774},
  year={2023}
}

@misc{qwen25vl,
      title={Qwen2.5-VL Technical Report}, 
      author={Shuai Bai and Keqin Chen and Xuejing Liu and Jialin Wang and Wenbin Ge and Sibo Song and Kai Dang and Peng Wang and Shijie Wang and Jun Tang and Humen Zhong and Yuanzhi Zhu and Mingkun Yang and Zhaohai Li and Jianqiang Wan and Pengfei Wang and Wei Ding and Zheren Fu and Yiheng Xu and Jiabo Ye and Xi Zhang and Tianbao Xie and Zesen Cheng and Hang Zhang and Zhibo Yang and Haiyang Xu and Junyang Lin},
      year={2025},
      eprint={2502.13923},
      archivePrefix={arXiv},
      primaryClass={cs.CV},
      url={https://arxiv.org/abs/2502.13923}, 
}

@article{qwen3vl,
  title={Qwen3-vl technical report},
  author={Bai, Shuai and Cai, Yuxuan and Chen, Ruizhe and Chen, Keqin and Chen, Xionghui and Cheng, Zesen and Deng, Lianghao and Ding, Wei and Gao, Chang and Ge, Chunjiang and others},
  journal={arXiv preprint arXiv:2511.21631},
  year={2025}
}

@inproceedings{dispider,
  title={Dispider: Enabling video llms with active real-time interaction via disentangled perception, decision, and reaction},
  author={Qian, Rui and Ding, Shuangrui and Dong, Xiaoyi and Zhang, Pan and Zang, Yuhang and Cao, Yuhang and Lin, Dahua and Wang, Jiaqi},
  booktitle={Proceedings of the Computer Vision and Pattern Recognition Conference},
  pages={24045--24055},
  year={2025}
}

@inproceedings{ovobench,
  title={Ovo-bench: How far is your video-llms from real-world online video understanding?},
  author={Niu, Junbo and Li, Yifei and Miao, Ziyang and Ge, Chunjiang and Zhou, Yuanhang and He, Qihao and Dong, Xiaoyi and Duan, Haodong and Ding, Shuangrui and Qian, Rui and others},
  booktitle={Proceedings of the Computer Vision and Pattern Recognition Conference},
  pages={18902--18913},
  year={2025}
}

@inproceedings{streamingbench,
  title={Streamingbench: Assessing the gap for mllms to achieve streaming video understanding},
  author={Lin, Junming and Fang, Zheng and Chen, Chi and Cheng, Haoxuan and Wan, Zihao and Luo, Fuwen and Wang, Ziyue and Li, Peng and Liu, Yang and Sun, Maosong},
  booktitle={ICASSP 2026-2026 IEEE International Conference on Acoustics, Speech and Signal Processing (ICASSP)},
  pages={12147--12151},
  year={2026},
  organization={IEEE}
}

@article{proactivevideoqa,
  title={Proactivevideoqa: A comprehensive benchmark evaluating proactive interactions in video large language models},
  author={Wang, Yueqian and Meng, Xiaojun and Wang, Yifan and Zhang, Huishuai and Zhao, Dongyan},
  journal={arXiv preprint arXiv:2507.09313},
  year={2025}
}

@inproceedings{ominimmi,
  title={Omnimmi: A comprehensive multi-modal interaction benchmark in streaming video contexts},
  author={Wang, Yuxuan and Wang, Yueqian and Chen, Bo and Wu, Tong and Zhao, Dongyan and Zheng, Zilong},
  booktitle={Proceedings of the IEEE/CVF Conference on Computer Vision and Pattern Recognition},
  pages={18925--18935},
  year={2025}
}

@article{etbench,
  title={Et bench: Towards open-ended event-level video-language understanding},
  author={Liu, Ye and Ma, Zongyang and Qi, Zhongang and Wu, Yang and Shan, Ying and Chen, Chang W},
  journal={Advances in Neural Information Processing Systems},
  volume={37},
  pages={32076--32110},
  year={2024}
}

@article{llava-video,
  title={Llava-video: Video instruction tuning with synthetic data},
  author={Zhang, Yuanhan and Wu, Jinming and Li, Wei and Li, Bo and Ma, Zejun and Liu, Ziwei and Li, Chunyuan},
  journal={arXiv preprint arXiv:2410.02713},
  year={2024}
}

@inproceedings{msad,
title = {Advancing Video Anomaly Detection: A Concise Review and a New Dataset},
author = {Liyun Zhu and Lei Wang and Arjun Raj and Tom Gedeon and Chen Chen},
booktitle = {The Thirty-eighth Conference on Neural Information Processing Systems Datasets and Benchmarks Track},
year = {2024}
}

@article{videochat-flash,
  title={Videochat-flash: Hierarchical compression for long-context video modeling},
  author={Li, Xinhao and Wang, Yi and Yu, Jiashuo and Zeng, Xiangyu and Zhu, Yuhan and Huang, Haian and Gao, Jianfei and Li, Kunchang and He, Yinan and Wang, Chenting and others},
  journal={arXiv preprint arXiv:2501.00574},
  year={2024}
}

@article{streamingvlm,
  title={Streamingvlm: Real-time understanding for infinite video streams},
  author={Xu, Ruyi and Xiao, Guangxuan and Chen, Yukang and He, Liuning and Peng, Kelly and Lu, Yao and Han, Song},
  journal={arXiv preprint arXiv:2510.09608},
  year={2025}
}

@inproceedings{cb,
  title={Class-balanced loss based on effective number of samples},
  author={Cui, Yin and Jia, Menglin and Lin, Tsung-Yi and Song, Yang and Belongie, Serge},
  booktitle={Proceedings of the IEEE/CVF conference on computer vision and pattern recognition},
  pages={9268--9277},
  year={2019}
}

@article{gemini1.5,
  title={Gemini 1.5: Unlocking multimodal understanding across millions of tokens of context},
  author={Team, Gemini and Georgiev, Petko and Lei, Ving Ian and Burnell, Ryan and Bai, Libin and Gulati, Anmol and Tanzer, Garrett and Vincent, Damien and Pan, Zhufeng and Wang, Shibo and others},
  journal={arXiv preprint arXiv:2403.05530},
  year={2024}
}

@article{grpo,
  title={Deepseek-r1: Incentivizing reasoning capability in llms via reinforcement learning},
  author={Guo, Daya and Yang, Dejian and Zhang, Haowei and Song, Junxiao and Wang, Peiyi and Zhu, Qihao and Xu, Runxin and Zhang, Ruoyu and Ma, Shirong and Bi, Xiao and others},
  journal={arXiv preprint arXiv:2501.12948},
  year={2025}
}

@inproceedings{videomme,
  title={Video-mme: The first-ever comprehensive evaluation benchmark of multi-modal llms in video analysis},
  author={Fu, Chaoyou and Dai, Yuhan and Luo, Yongdong and Li, Lei and Ren, Shuhuai and Zhang, Renrui and Wang, Zihan and Zhou, Chenyu and Shen, Yunhang and Zhang, Mengdan and others},
  booktitle={Proceedings of the IEEE/CVF conference on computer vision and pattern recognition},
  pages={24108--24118},
  year={2025}
}

@article{longvideobench,
  title={Longvideobench: A benchmark for long-context interleaved video-language understanding},
  author={Wu, Haoning and Li, Dongxu and Chen, Bei and Li, Junnan},
  journal={Advances in Neural Information Processing Systems},
  volume={37},
  pages={28828--28857},
  year={2024}
}

@article{thinking-qwenvl,
  title={Progressive Online Video Understanding with Evidence-Aligned Timing and Transparent Decisions},
  author={Zhang, Kecheng and Yang, Zongxin and Han, Mingfei and Hao, Haihong and Zhuge, Yunzhi and Li, Changlin and Zhao, Junhan and Li, Zhihui and Chang, Xiaojun},
  journal={arXiv preprint arXiv:2604.18459},
  year={2026}
}

@inproceedings{querystream,
  title={QueryStream: Advancing Streaming Video Understanding with Query-Aware Pruning and Proactive Response},
  author={Zhang, Kairui and Yang, Zhenyu and Wang, Bing and Qian, Shengsheng and Xu, Changsheng},
  booktitle={The Fourteenth International Conference on Learning Representations},
  year={2026}
}

@misc{minicpm_o_45,
  title        = {MiniCPM-o 4.5 Technical Report},
  author       = {OpenBMB},
  year         = {2026},
  howpublished = {\url{https://github.com/OpenBMB/MiniCPM-o/blob/main/docs/MiniCPM_o_45_technical_report.pdf}},
  note         = {GitHub technical report}
}

@inproceedings{streammind,
  title={Streammind: Unlocking full frame rate streaming video dialogue through event-gated cognition},
  author={Ding, Xin and Wu, Hao and Yang, Yifan and Jiang, Shiqi and Zhang, Qianxi and Bai, Donglin and Chen, Zhibo and Cao, Ting},
  booktitle={Proceedings of the IEEE/CVF International Conference on Computer Vision},
  pages={13448--13459},
  year={2025}
}

@article{livestar,
  title={LiveStar: Live Streaming Assistant for Real-World Online Video Understanding},
  author={Yang, Zhenyu and Zhang, Kairui and Hu, Yuhang and Wang, Bing and Qian, Shengsheng and Wen, Bin and Yang, Fan and Gao, Tingting and Dong, Weiming and Xu, Changsheng},
  journal={arXiv preprint arXiv:2511.05299},
  year={2025}
}

@article{thinkstream,
  title={Thinking in Streaming Video},
  author={Liu, Zikang and Guo, Longteng and Li, Handong and Zhen, Ru and He, Xingjian and Ji, Ruyi and Ren, Xiaoming and Zhang, Yanhao and Lu, Haonan and Liu, Jing},
  journal={arXiv preprint arXiv:2603.12938},
  year={2026}
}

@misc{tom,
  title        = {Learning to Respond: A Large-Scale Benchmark and Progressive Learning Framework for Trigger-Centric Online Video Understanding},
  author       = {Qian, Jiawen and Du, Hang and Nan, Guoshun and Huang, Shan and Yu, Jiaqi and Wang, Haoqing and Chen, Jingfeng and Cai, Mengxue and Yang, Mengdi and Li, Jiarong and Li, Zhilong and Wang, Hua and Liu, Jun and Jiang, Xudong and Leng, Sicong},
  year         = {2025},
  howpublished = {\url{https://openreview.net/pdf?id=gmpnSSiJt7}},
}

@article{llava-onevision,
  title={Llava-onevision: Easy visual task transfer},
  author={Li, Bo and Zhang, Yuanhan and Guo, Dong and Zhang, Renrui and Li, Feng and Zhang, Hao and Zhang, Kaichen and Zhang, Peiyuan and Li, Yanwei and Liu, Ziwei and others},
  journal={arXiv preprint arXiv:2408.03326},
  year={2024}
}

@article{qwen2vl,
  title={Qwen2-vl: Enhancing vision-language model's perception of the world at any resolution},
  author={Wang, Peng and Bai, Shuai and Tan, Sinan and Wang, Shijie and Fan, Zhihao and Bai, Jinze and Chen, Keqin and Liu, Xuejing and Wang, Jialin and Ge, Wenbin and others},
  journal={arXiv preprint arXiv:2409.12191},
  year={2024}
}

@article{internvl3,
  title={Internvl3. 5: Advancing open-source multimodal models in versatility, reasoning, and efficiency},
  author={Wang, Weiyun and Gao, Zhangwei and Gu, Lixin and Pu, Hengjun and Cui, Long and Wei, Xingguang and Liu, Zhaoyang and Jing, Linglin and Ye, Shenglong and Shao, Jie and others},
  journal={arXiv preprint arXiv:2508.18265},
  year={2025}
}

@article{aura,
  title={AURA: Always-On Understanding and Real-Time Assistance via Video Streams},
  author={Lu, Xudong and Bo, Yang and Chen, Jinpeng and Li, Shuhan and Guo, Xintong and Guan, Huankang and Liu, Fang and Xu, Dunyuan and Sun, Peiwen and Sun, Heyang and others},
  journal={arXiv preprint arXiv:2604.04184},
  year={2026}
}

@article{chiang2024chatbot,
  title={Chatbot arena: An open platform for evaluating llms by human preference},
  author={Chiang, Wei-Lin and Zheng, Lianmin and Sheng, Ying and Angelopoulos, Anastasios Nikolas and Li, Tianle and Li, Dacheng and Zhang, Hao and Zhu, Banghua and Jordan, Michael and Gonzalez, Joseph E and others},
  journal={arXiv preprint arXiv:2403.04132},
  year={2024}
}

@book{efron1994introduction,
  title={An introduction to the bootstrap},
  author={Efron, Bradley and Tibshirani, Robert J},
  year={1994},
  publisher={Chapman and Hall/CRC}
}

@article{bradley1952rank,
  title={Rank analysis of incomplete block designs: I. the method of paired comparisons},
  author={Bradley, Ralph Allan and Terry, Milton E},
  journal={Biometrika},
  volume={39},
  number={3/4},
  pages={324--345},
  year={1952},
  publisher={JSTOR}
}
}

\clearpage
\appendix
\section{Data Construction}
\label{sec:data_construction}

\subsection{Data Construction Pipeline}
\label{subsec:data_construction_pipeline}
We construct StreamPro-Bench, StreamPro-SFT-63K, and StreamPro-RL-3K through a pipeline of video collection and filtering, data generation, and verification.

\noindent\textbf{Video Collection and Filtering}
We collect raw videos from four sources: ET-Instruct-164k~\citep{etbench}, LLaVA-Video-178k~\citep{llava-video}, MSAD~\citep{msad}, and manually crawled videos from diverse online platforms. 
To ensure sufficient temporal dynamics for proactive QA, we employ Qwen3-VL-8B-Instruct~\citep{qwen3vl} to retain videos that satisfy two criteria: (1) the number of scene transitions exceeds a predefined threshold, and (2) each scene exhibits a sufficiently long duration. 
This pre-filtering stage removes static or overly monotonous videos that are unsuitable for proactive QA tasks. 
For task-specific quality control, we further apply targeted filtering strategies: for Procedural Guidance, we enforce \emph{step uniqueness} to ensure that procedural steps are clearly distinguishable and non-redundant; for Risk Forecasting, we apply an \emph{egocentric + outdoor} filter to retain first-person outdoor videos aligned with real-world assistive navigation scenarios.

\noindent\textbf{Data Generation}
For each filtered video, we construct three complementary types of captions: 
(1)~\textit{Event-level captions} that describe high-level activities and temporal events. (2)~\textit{Action-level captions} that focus on human actions and behaviors. (3)~\textit{Object-level captions} that enumerate salient objects and their attributes. 
Based on these multi-granularity captions, we synthesize QA pairs for most tasks. Risk Forecasting follows the manual annotation procedure in Section~\ref{subsec:risk_forecasting_annotation}.

\noindent\textbf{Data Verification}
Given that proactive streaming video understanding requires each question to be associated with multiple temporally grounded answers, we verify generated samples through the following steps.
\textit{(1) Basic Verification.} Each synthesized QA pair is first subjected to a basic verification and refinement process, where we correct semantic inconsistencies and align answers with their corresponding temporal segments.
\textit{(2) Two-Agent Iterative Verification.} A Discrimination Agent evaluates each QA pair against predefined quality criteria. If a sample is deemed unsatisfactory, it returns structured rejection feedback. A Generation Agent then revises the sample accordingly. This discriminate–generate loop is executed for up to three iterations to progressively improve data quality.
\textit{(3) Human Review.} StreamPro-Bench QA pairs that pass the two-agent verification stage undergo final human review to ensure correctness and consistency. All steps involving LLMs are implemented using Gemini 2.5 Pro~\citep{gemini2.5}.
For the training datasets, we omit the human review step and further filter the generated samples to obtain StreamPro-SFT-63K and StreamPro-RL-3K.

\subsection{Annotation Guideline for Risk Forecasting}
\label{subsec:risk_forecasting_annotation}
As for Risk Forecasting task, we find it is too difficult to generate reliable annotation automatically. Thus, we adopt a human annotation pipeline and human review to ensure its quality.

Annotators are instructed to estimate when a potential accident would occur under the assumption of constant pedestrian velocity and trajectory. 
The warning moment is defined as \emph{3 seconds before the estimated accident time}, and annotators are required to mark this moment accordingly.
After initial annotation, all labels are further reviewed and refined by two independent human to ensure annotation quality.

\noindent\textbf{Risk Taxonomy.} We define four categories of risks to systematically characterize navigation hazards in streaming egocentric video.
\begin{enumerate}
    \item \textbf{Structural Terrain}, referring to ground-level geometric irregularities that may compromise pedestrian stability (e.g., stairs, curbs, slopes).
    \item \textbf{Path Blockage}, denoting physical obstacles that directly obstruct the walking trajectory (e.g., parked bicycles, barriers).
    \item \textbf{Overhanging Objects}, representing elevated hazards that are difficult to detect using a white cane due to their height (e.g., low branches, signboards).
    \item \textbf{Functional Elements}, capturing semantically critical environmental cues necessary for navigation decisions (e.g., traffic lights, crosswalks).
\end{enumerate}

\subsection{StreamPro-SFT-63K and StreamPro-RL-3K}
\label{subsec:streampro_training_data}

\begin{figure}[ht]
  \centering

  \begin{subfigure}[b]{0.48\columnwidth}
    \centering
    \includegraphics[height=4cm, width=\textwidth, keepaspectratio]{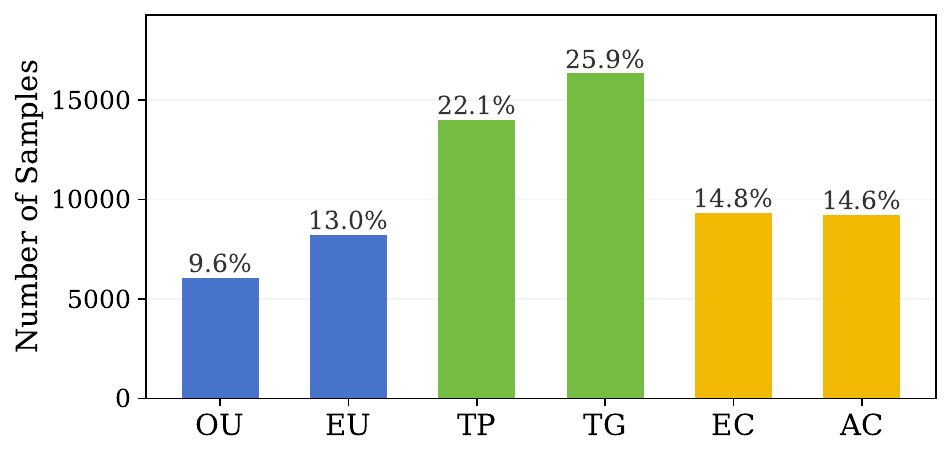}
    \label{fig:sft_data}
  \end{subfigure}
  \hfill
  \begin{subfigure}[b]{0.48\columnwidth}
    \centering
    \includegraphics[height=4cm, width=\textwidth, keepaspectratio]{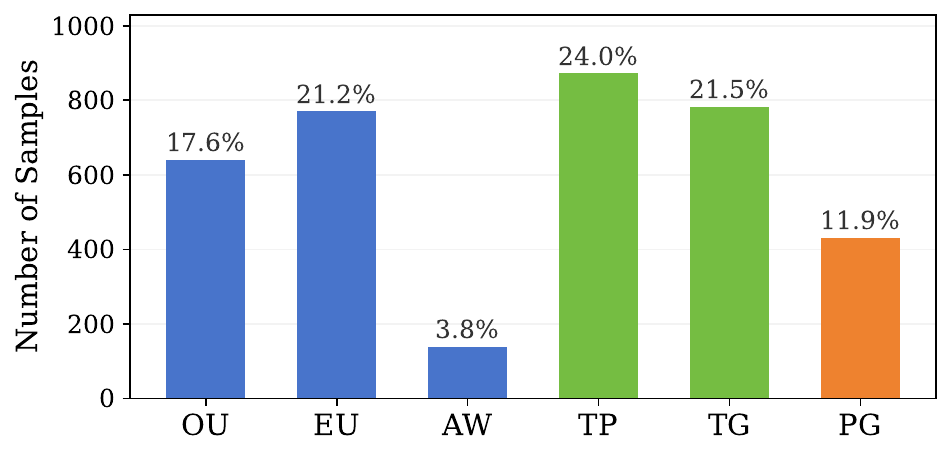}
    \label{fig:rl_data}
  \end{subfigure}
  \caption{Dataset Statistics. \textbf{Left}: StreamPro-SFT-63K; \textbf{Right}: StreamPro-RL-3K.
  }
  \label{fig:data_distribution}
\end{figure}

In the SFT stage, our primary objective is to endow the model with fundamental proactive capabilities. 
The training tasks encompass Event Captioning (EC), Action Captioning (AC), Event Understanding (EU), Object Understanding (OU), Temporal Perception (TP), and Temporal Grounding (TG).
In the RL stage, we further introduce more challenging proactive tasks.
The training tasks encompass Event Understanding (EU), Object Understanding (OU), Anomaly Warning (AW), Temporal Perception (TP), Temporal Grounding (TG) and Procedural Guidance (PG). Neither training dataset includes Risk Forecasting samples due to the difficulty of constructing this task.
The task distributions of the two datasets are shown in Figure~\ref{fig:data_distribution}.

\section{StreamPro-Bench Evaluation}
\label{sec:streampro_bench_evaluation}

\subsection{Offline Model Evaluation}
\label{subsec:offline_model_evaluation}

For offline models, we evaluate Qwen2.5-VL-7B-Instruct~\citep{qwen25vl} and Qwen3-VL-8B-Instruct~\citep{qwen3vl} with a prompt-based decision protocol designed to simulate proactive responses to a video stream. The evaluation prompt is provided in Section~\ref{sec:prompt_templates}. Video frames are sampled at 1 frame per second by default and processed sequentially. After the query is issued, the model generates a response for each sampled frame using the retained conversation history, which is limited to the most recent 200 frame turns by default. The prompt instructs the model to output “Wait” when the available visual context is insufficient to answer without guessing. When the context is sufficient, it instructs the model to output “Wait” if the inferred answer is unchanged from its previous answer, and otherwise to output the updated answer. Each decision therefore uses only the frames available up to the current time. In this way, an offline can be evaluated in a streaming context reasonably.

\subsection{Task-Specific Temporal Tolerance Windows}
\label{subsec:temporal_tolerance_windows}

\begin{table}[t]
    \centering
    \small

    \setlength{\tabcolsep}{3pt}
    \caption{Optimal response times and evaluation windows across tasks. $t_{\mathrm{opt}}$ denotes the optimal response time. For Event Understanding and Temporal Perception, $t_{\mathrm{start}}$ and $t_{\mathrm{end}}$ denote the event boundaries.}
    \label{tab:time_window_resize}
    \adjustbox{max width=\linewidth}{
    \begin{tabular}{lll}
        \toprule

        \textbf{Task} &
        \textbf{Optimal Response Time} &
        \textbf{Evaluation Window} \\

        \midrule

        EU&
        Within event duration &
        $[t_{\mathrm{start}},\,t_{\mathrm{end}}+3\mathrm{s}]$ \\

        OU&
        Object state transition moment &
        $[t_{\mathrm{opt}}-3\mathrm{s},\,t_{\mathrm{opt}}+3\mathrm{s}]$ \\

        AW&
        Anomaly onset time &
        $[t_{\mathrm{opt}},\,t_{\mathrm{opt}}+5\mathrm{s}]$ \\

        TP&
        Within event duration &
        $[t_{\mathrm{start}},\,t_{\mathrm{end}}+3\mathrm{s}]$ \\

        TG&
        Event ending time &
        $[t_{\mathrm{opt}}-3\mathrm{s},\,t_{\mathrm{opt}}+3\mathrm{s}]$ \\

        PG&
        Previous step completion time &
        $[t_{\mathrm{opt}}-3\mathrm{s},\,t_{\mathrm{opt}}]$ \\

        RF&
        3s before the estimated accident time &
        $[t_{\mathrm{opt}}-1\mathrm{s},\,t_{\mathrm{opt}}+3\mathrm{s}]$ \\

        \bottomrule
    \end{tabular}
    }
\end{table}

To accommodate the heterogeneous temporal characteristics of different tasks, we define task-specific optimal response timings and corresponding temporal tolerance windows, as summarized in Table~\ref{tab:time_window_resize}. 
Event Understanding and Temporal Perception use the event interval as the optimum. For these tasks, $\operatorname{dist}(t_{\mathrm{pred}},t_{\mathrm{gt}})$ in the timing score of Section~\ref*{main:subsec:eval_metric} is zero within the interval and otherwise equals the distance to its nearest boundary.
For the remaining tasks, the optimum is a timestamp and the distance is the absolute difference from it.
Only responses within the listed matching windows are eligible.
We use $\tau=4$ for all tasks except Anomaly Warning, where $\tau=6$, and Risk Forecasting, where $\tau=2$ before the optimal warning time and $\tau=4$ after it.

\subsection{Human Preference Alignment Validation}
\label{subsec:human_preference_alignment}

To validate the proposed StreamPro-F1 metric and demonstrate its alignment with human judgment, we conduct a pairwise human evaluation.

\begin{figure*}[t]
    \centering
    \begin{minipage}[b]{0.36\textwidth}
        \centering
        \includegraphics[width=\linewidth]{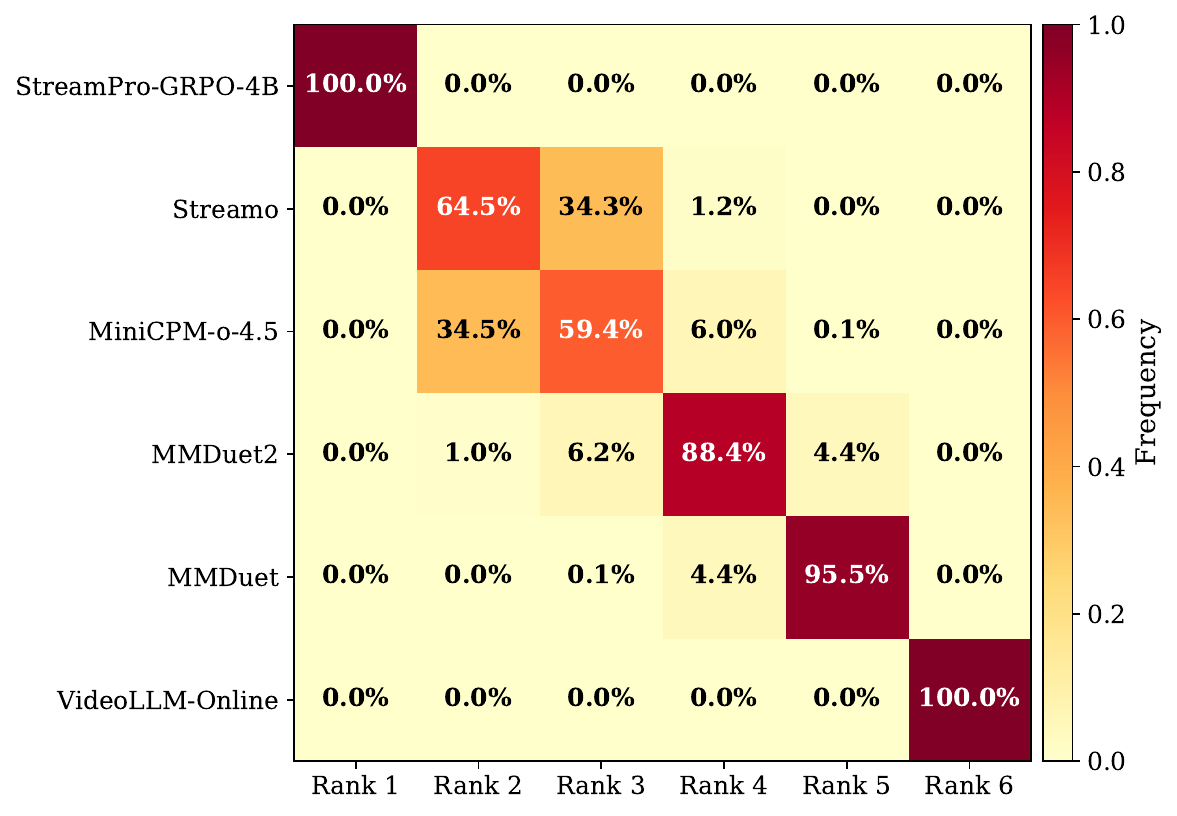}
    \end{minipage}\hfill
    \begin{minipage}[b]{0.62\textwidth}
        \centering
        \includegraphics[width=\linewidth]{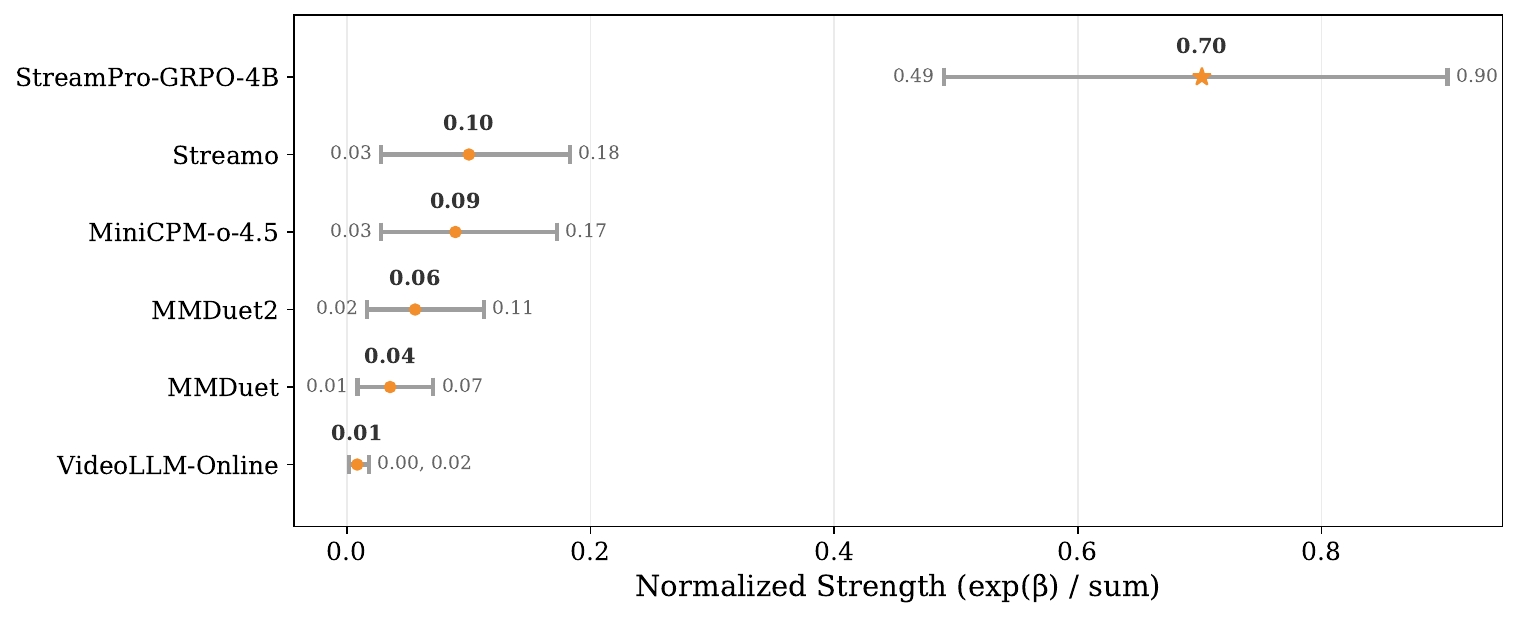}
    \end{minipage}
    \caption{Human preference ranking reliability across 1,000 bootstrap samples. The left panel shows rank stability, and the right panel shows Bradley-Terry scores with 95\% confidence intervals.}
    \label{fig:human_rank_reliability}
\end{figure*}

\noindent\textbf{Evaluation Methodology.}
Following the Chatbot Arena paradigm~\citep{chiang2024chatbot}, human annotators evaluate anonymized response trajectories from two sampled models and declare a winner or a tie. 
To infer global rankings from these relative preferences, we employ the Bradley-Terry (BT) model~\citep{bradley1952rank}. 
Given the inherent difficulty of streaming tasks, ties are frequent. We handle them via the Weighted BT approach~\citep{chiang2024chatbot}, encoding each tie as two symmetric $0.5$-weighted pseudo-observations, allowing seamless likelihood optimization without discarding data.

\noindent\textbf{Ranking Reliability Assessment.}
To quantify the statistical reliability of the derived rankings, we employ Bootstrap resampling~\citep{efron1994introduction}. 
Specifically, we generate $1{,}000$ bootstrap samples with replacement from the collected pairwise comparisons and independently fit the Weighted BT model on each. 
This yields a Rank Stability Matrix (frequency of a model occupying each rank) and Score Confidence Intervals (median BT parameter with $95\%$ CI).

\noindent\textbf{Results and Analysis.}
The rank stability heatmap and score confidence intervals in Figure~\ref{fig:human_rank_reliability} show highly confident stratification among the evaluated models. 
Our proposed StreamPro-GRPO-4B unambiguously secures the top rank across all bootstrap replicates. Its confidence bounds are completely detached from the rest of the cohort, demonstrating absolute stability and significant superiority in proactive streaming capabilities. 
While minor variance exists among closely matched middle-tier models (e.g., Streamo and MiniCPM-o-4.5), the hierarchical boundaries remain strictly defined, with VideoLLM-Online consistently anchoring the final rank.

\noindent\textbf{Alignment with Benchmark Metric.}
Crucially, the final human preference ranking perfectly matches the ranking produced by our proposed StreamPro-F1 metric (yielding a Spearman's rank correlation $\rho = 1.0$). 
This absolute consensus demonstrates that our automated evaluation protocol faithfully captures human perception of proactive streaming quality, validating StreamPro-Bench as a robust and reliable evaluation standard.

\section{Additional Experiments}
\label{sec:additional_experiments}


\subsection{Matching Strategies}
\label{subsec:matching_strategies}

Here we describe how predicted responses are matched to ground truth responses in StreamPro-F1 and the RM reward.

\noindent\textbf{Matching Strategy for StreamPro-F1 (Prediction-First).}
A valid prediction response is the content following \texttt{</Response>} that is not empty. We process valid predicted responses in chronological order. For each prediction, we scan the ground truth in their chronological order and assign the prediction to the first ground truth whose temporal tolerance window contains the prediction response time. Each prediction is assigned to at most one ground truth. If multiple predictions are assigned to the same ground truth, only the prediction with the highest joint score $S_{ij}=S_{\mathrm{time},ij}S_{\mathrm{sem},ij}$ contributes to StreamPro-F1. All valid predictions remain in the precision denominator, including those outside every evaluation window and those not selected for scoring. Code~\ref{lst:streampro_f1_matching} shows the matching step. As the outer loop iterates over predictions, we name this strategy \textit{prediction-first} matching.

\begin{codebox}[label={lst:streampro_f1_matching}]{Python}{StreamPro-F1 matching}
best = [0.0] * len(ground_truth)
for pred in valid_preds:
    for i, gt in enumerate(ground_truth):
        if in_window(pred, gt, task):
            score = joint_score(pred, gt)
            best[i] = max(best[i], score)
            break
\end{codebox}

\noindent\textbf{Matching Strategy for the RM Reward (Ground-Truth-First).}
For the RM reward, we process ground truth in chronological order. For each ground truth, we consider valid prediction responses that have not been selected previously and whose response time falls in the temporal tolerance window of the ground truth under the tolerance $\tau$. We select the candidate with the highest pair score
$S'_{ij}=(S_{\mathrm{time},ij}+S_{\mathrm{sem},ij})/2$
and remove the prediction from subsequent candidate sets. Predictions are not assigned to the first eligible ground truth window before their pair scores are compared. Selected pair scores contribute to Equation~\ref*{main:eq:rm_reward}. Unmatched predictions and ground truth remain in its denominator. In Code~\ref{lst:rm_matching}, \texttt{preds} denotes these outputs and \texttt{PG} denotes Procedural Guidance. The implementation returns a reward of $1$ when both the prediction and ground truth sets are empty. As the outer loop iterates over ground truth, we name this strategy \textit{ground-truth-first} matching.

\begin{codebox}[label={lst:rm_matching}]{Python}{RM reward matching}
unmatched_preds = set(range(len(preds)))
matches = []
for gt in ground_truth:
    candidates = []
    for i in sorted(unmatched_preds):
        pred = preds[i]
        if timing(pred, gt, tau) <= 0:
            continue
        if task == "PG":
            if pred.time > gt.time:
                continue
        candidates.append(i)
    if candidates:
        i = max(
            candidates,
            key=lambda k: rm_score(
                preds[k], gt
            )
        )
        matches.append((i, gt))
        unmatched_preds.remove(i)
\end{codebox}

\noindent\textbf{Impact of Matching Strategy in RL.}
We also ablate the matching strategy used to assign predictions to ground truth during RL. 
We compare \textit{ground-truth-first} and \textit{prediction-first} matching in the RL stage.
As shown in Table~\ref{tab:ablation_matching}, ground-truth-first matching improves SPB from 20.7 to 25.1, OVO-RTVP from 63.8 to 64.2, and VideoMME from 60.3 to 60.9.
Prediction-first uses the same assignment order as benchmark evaluation, but assigns each prediction to the first eligible ground truth before comparing pair scores.
When evaluation windows overlap, this assignment may prevent a prediction from contributing to a later ground truth with which it has a higher pair score.
Ground-truth-first instead selects the highest-scoring available prediction for each ground truth, which can provide more informative reward for responses near overlapping windows.


\begin{table}[t]
    \centering
    \small
    \caption{Ablation of RM matching strategies during GRPO.}
    \label{tab:ablation_matching}
    \begin{adjustbox}{max width=\linewidth}
    \begin{tabular}{l ccc}
        \toprule
        \textbf{Matching Strategy} & \textbf{SPB} & \textbf{OVO-RTVP} & \textbf{VideoMME} \\
        \midrule
        Prediction-first & 20.7 & 63.8 & 60.3 \\
        \rowcolor{our_color!10}
        GT-first & \textbf{25.1} & \textbf{64.2} & \textbf{60.9} \\
        \bottomrule
    \end{tabular}
    \end{adjustbox}
\end{table}

\subsection{StreamPro-F1 across RL training steps.}
\label{subsec:rl_training_curves}

\begin{figure}[t]
    \centering
    \includegraphics[width=0.9\linewidth]{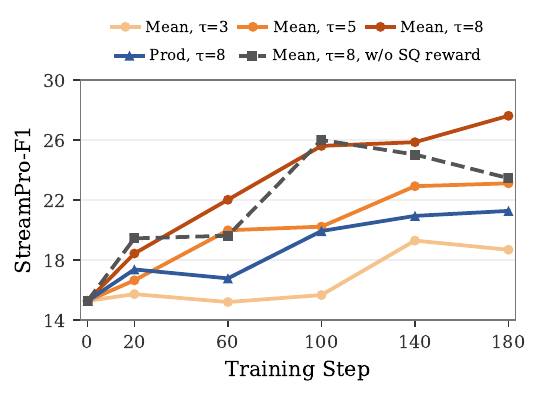}
    \caption{StreamPro-F1 during RL training on a subset of StreamPro-Bench.}
    \label{fig:rl_training_curves}
\end{figure}

Figure~\ref{fig:rl_training_curves} complements the reward ablation in Table~\ref*{main:tab:ablation_rl} by tracking StreamPro-F1 during RL training on a subset of StreamPro-Bench. All configurations start from a similar score, but their trajectories diverge as training proceeds. With the mean RM score and SQ reward, $\tau=8$ improves steadily from the early steps and continues to gain after step 140. The $\tau=5$ curve rises more slowly and levels off between steps 60 and 100 before improving again, while $\tau=3$ changes little through step 100. These trends suggest that a wider temporal tolerance provides useful reward signals earlier in training.

At $\tau=8$, product scoring dips between steps 20 and 60 and recovers more slowly than mean scoring. Removing the SQ reward produces a rapid early gain and briefly gives the highest score around step 100, but the curve then declines. The configuration with SQ maintains its gains and improves further toward the end of training. This pattern suggests that SQ helps sustain progress over longer training, rather than simply accelerating the initial improvement.

\section{Detailed Benchmark Results}
\label{sec:detailed_benchmark_results}

\noindent\textbf{Results on StreamPro-Bench.}
As shown in Table~\ref{tab:perceptual_understanding}, Table~\ref{tab:temporal_reasoning} and Table~\ref{tab:anticipatory_assistance}, we present comprehensive results on StreamPro-Bench across all three evaluation dimensions and seven tasks, reporting Time Score, Semantic Score, Precision, Recall, and StreamPro-F1, where higher values indicate better performance.

\emph{Existing proactive models perform poorly on StreamPro-Bench, indicating that their capabilities are still far from those required for an ideal real-world proactive system.} 
Among the three evaluated dimensions, they achieve relatively acceptable performance in perceptual understanding, while showing significant deficiencies in temporal reasoning and anticipatory assistance.
Specifically, compared with other proactive models, MMDuet~\citep{mmduet} and MMDuet2~\citep{mmduet2} achieve relatively high Time Score, Semantic Score, and Recall, since they tend to generate responses more frequently, which increases the probability of hitting the correct temporal window. 
However, due to this tendency toward over-generation, Precision is penalized and thus drops to a more moderate level. 
Besides, Streamo~\citep{streamo} demonstrates strong temporal reasoning capabilities, due to the presence of video temporal grounding tasks in its training data.  
Note that VideoLLM-Online~\citep{videollmonline} exhibits an inflated performance on Procedural Guidance, which is mainly caused by its tendency to produce overly generic and highly abstract responses. Such responses are often assigned relatively stable and moderate scores by the LLM-based evaluator across different cases, leading to an apparently high average score. 

\noindent\textbf{Results on StreamingBench and OVO-Bench.}
Table~\ref{tab:streaming_bench_full} and Table~\ref{tab:ovo_bench_full} present the detailed results for RTVU and CU task in StreamingBench, and RTVP in OVO-Bench, respectively.

\begin{table*}[p]
    \centering
    \small
    \renewcommand{\arraystretch}{1.2}
    \setlength{\tabcolsep}{3pt}
    \caption{Detailed Results on Perceptual Understanding.}
    \label{tab:perceptual_understanding}
    \begin{adjustbox}{max width=\linewidth}
    \begin{tabular}{l c ccccc ccccc ccccc c}
        \toprule
        \multirow{2}{*}{\textbf{Methods}} & \multirow{2}{*}{\textbf{Params}} 
        & \multicolumn{5}{c}{\textbf{Event Understanding}} 
        & \multicolumn{5}{c}{\textbf{Object Understanding}} 
        & \multicolumn{5}{c}{\textbf{Anomaly Warning}} 
        & \multirow{2}{*}{\textbf{Avg. F1}} \\
        \cmidrule(lr){3-7} \cmidrule(lr){8-12} \cmidrule(lr){13-17}
        & & Time & Sem. & P & R & F1 
          & Time & Sem. & P & R & F1 
          & Time & Sem. & P & R & F1 
          & \\
        \midrule
        \multicolumn{18}{c}{\textit{Offline Models + Proactive Prompt}} \\
        \midrule
        Qwen2.5-VL-7B~\citep{qwen25vl}  & 7B 
        & 8.9 & 8.0 & 0.3 & 4.6 & 0.5
        & 15.6 & 19.2 & 19.8 & 11.7 & 11.3
        & 1.7 & 21.7 & 0.7 & 1.4 & 0.8
        & 4.2 \\

        Qwen3-VL-8B~\citep{qwen3vl}    & 8B 
        & 43.1 & 25.1 & 1.7 & 24.2 & 2.5
        & 68.9 & 61.8 & 1.7 & 57.8 & 3.0
        & 66.6 & 63.3 & 7.2 & 48.7 & 11.7
        & 5.7 \\

        \midrule
        \multicolumn{18}{c}{\textit{Open-Source Proactive Models}} \\
        \midrule
        VideoLLM-Online~\citep{videollmonline} & 8B 
        & 0.6 & 0.1 & 0.3 & 0.1 & 0.1
        & 11.4 & 0.0 & 0.0 & 0.0 & 0.0
        & 31.7 & 3.7 & 0.7 & 2.6 & 0.8
        & 0.3 \\
        
        MMDuet~\citep{mmduet}          & 7B 
        & 82.1 & 39.4 & 8.4 & 36.0 & 13.0
        & 44.0 & 40.7 & 2.7 & 23.2 & 4.8
        & 25.5 & 24.1 & 9.1 & 10.4 & 9.4
        & 9.1 \\
        
        MMDuet2~\citep{mmduet2}         & 3B 
        & 81.6 & 59.3 & 11.8 & 56.5 & 17.7
        & 31.5 & 9.5 & 3.1 & 4.9 & 3.3
        & 9.9 & 18.7 & 7.9 & 8.7 & 8.1
        & 9.7 \\

        MiniCPM-o-4.5~\citep{minicpm_o_45}   & 9B 
        & 10.9 & 8.0 & 15.5 & 7.7 & 9.8
        & 14.4 & 20.9 & 25.9 & 9.6 & 13.9
        & 31.0 & 21.7 & 15.3 & 16.9 & 15.8
        & 13.2 \\

        Streamo~\citep{streamo}         & 3B 
        & 13.4 & 8.2 & 20.5 & 7.4 & 8.9
        & 57.9 & 49.5 & 12.5 & 40.8 & 15.8
        & 2.6 & 3.9 & 2.3 & 2.3 & 2.3
        & 9.0 \\

        \rowcolor{our_color!10}
        \textbf{StreamPro-SFT}  & \textbf{3B} 
        & 86.1 & 58.3 & 11.9 & 56.1 & 18.4
        & 48.6 & 54.2 & 8.1 & 41.6 & 12.2
        & 20.7 & 19.8 & 9.9 & 12.4 & 10.5
        & 13.7 \\
        
        \rowcolor{our_color!10}
        \textbf{StreamPro-GRPO} & \textbf{3B} 
        & 75.3 & 57.0 & 42.6 & 53.6 & \underline{46.1} & 59.1 & 66.7 & 36.5 & 50.9 & \textbf{41.0} & 32.5 & 30.2 & 17.7 & 18.5 & 18.0 & \underline{35.0} \\
        
        \rowcolor{our_color!20}
        \textbf{StreamPro-SFT}  & \textbf{4B} 
        & 90.2 & 67.3 & 18.0 & 65.2 & 26.3
        & 53.2 & 58.4 & 11.6 & 45.9 & 16.7
        & 58.3 & 59.8 & 24.8 & 39.6 & \underline{28.5}
        & 23.8 \\
        
        \rowcolor{our_color!20}
        \textbf{StreamPro-GRPO} & \textbf{4B} 
        & 82.0 & 67.4 & 49.2 & 64.6 & \textbf{54.6} & 51.9 & 49.5 & 35.1 & 40.1 & \underline{36.4} & 63.3 & 64.9 & 41.5 & 48.8 & \textbf{43.5} & \textbf{44.8} \\
        
        \bottomrule
    \end{tabular}
\end{adjustbox}
\end{table*}

\begin{table*}[t]
    \centering
    \small
    \renewcommand{\arraystretch}{1.2}
    \setlength{\tabcolsep}{3pt}
    \caption{Detailed Results on Temporal Reasoning.}
    \label{tab:temporal_reasoning}
    \begin{tabularx}{\linewidth}{l c *{11}{Y}}
        \toprule
        \multirow{2}{*}{\textbf{Methods}} & \multirow{2}{*}{\textbf{Params}} & \multicolumn{5}{c}{\textbf{Temporal Perception}} & \multicolumn{5}{c}{\textbf{Temporal Grounding}} & \multirow{2}{*}{\textbf{Avg. F1}} \\
        \cmidrule(lr){3-7} \cmidrule(lr){8-12}
        & & Time & Sem. & P & R & F1 & Time & Sem. & P & R & F1 & \\
        \midrule
        \multicolumn{13}{c}{\textit{Offline Models + Proactive Prompt}} \\
        \midrule
        Qwen2.5-VL-7B~\citep{qwen25vl}  & 7B 
        & 1.1 & 1.2 & 0.7 & 1.0 & 0.8 
        & 8.8 & 0.5 & 0.4 & 0.4 & 0.4 & 0.6 \\

        Qwen3-VL-8B~\citep{qwen3vl}    & 8B 
        & 50.8 & 34.8 & 0.9 & 34.3 & 1.6 
        & 6.8 & 0.3 & 0.1 & 0.2 & 0.1 & 0.9 \\

        \midrule
        \multicolumn{13}{c}{\textit{Open-Source Proactive Models}} \\
        \midrule
        VideoLLM-Online~\citep{videollmonline} & 8B 
        & 0.7 & 0.1 & 0.0 & 0.1 & 0.0 
        & 0.4 & 0.0 & 0.0 & 0.0 & 0.0 & 0.0 \\

        MMDuet~\citep{mmduet}          & 7B 
        & 91.8 & 40.6 & 1.6 & 38.6 & 3.0 
        & 53.9 & 2.1 & 0.1 & 1.1 & 0.1 & 1.6 \\

        MMDuet2~\citep{mmduet2}         & 3B 
        & 85.8 & 63.6 & 4.0 & 61.3 & 6.7 
        & 12.7 & 0.0 & 0.0 & 0.0 & 0.0 & 3.4 \\

        MiniCPM-o-4.5~\citep{minicpm_o_45}   & 9B 
        & 1.3 & 0.8 & 0.1 & 0.6 & 0.1 
        & 3.5 & 0.0 & 0.0 & 0.0 & 0.0 & 0.1 \\

        Streamo~\citep{streamo}         & 3B 
        & 39.2 & 26.6 & 18.4 & 25.4 & 20.4 
        & 16.5 & 12.4 & 8.1 & 8.6 & 8.2 & 14.3 \\

        \rowcolor{our_color!10}
        \textbf{StreamPro-SFT}  & \textbf{3B} 
        & 65.8 & 38.7 & 37.2 & 38.6 & 37.6
        & 25.8 & 14.9 & 5.4 & 11.3 & 6.4 & 22.0 \\
        
        \rowcolor{our_color!10}
        \textbf{StreamPro-GRPO} & \textbf{3B} 
        & 79.7 & 62.8 & 45.7 & 60.3 & 50.3 
        & 37.6 & 32.0 & 17.4 & 23.6 & \underline{19.2} & \underline{34.7} \\
        
        \rowcolor{our_color!20}
        \textbf{StreamPro-SFT}  & \textbf{4B} 
        & 69.4 & 51.4 & 50.6 & 51.3 & \underline{50.7}
        & 23.3 & 15.6 & 6.3 & 11.8 & 7.4 & 29.1 \\
        
        \rowcolor{our_color!20}
        \textbf{StreamPro-GRPO} & \textbf{4B} 
        & 83.3 & 71.6 & 54.8 & 69.0 & \textbf{58.8} 
        & 53.4 & 50.5 & 29.0 & 38.1 & \textbf{31.4} & \textbf{45.1} \\
        \bottomrule
    \end{tabularx}
\end{table*}

\begin{table*}[t]
    \centering
    \small
    \renewcommand{\arraystretch}{1.2}
    \setlength{\tabcolsep}{3pt}
    \caption{Detailed Results on Anticipatory Assistance.}
    \label{tab:anticipatory_assistance}
    \begin{tabularx}{\linewidth}{l c *{11}{Y}}
        \toprule
        \multirow{2}{*}{\textbf{Methods}} & \multirow{2}{*}{\textbf{Params}} & \multicolumn{5}{c}{\textbf{Procedural Guidance}} & \multicolumn{5}{c}{\textbf{Risk Forecasting}} & \multirow{2}{*}{\textbf{Avg. F1}} \\
        \cmidrule(lr){3-7} \cmidrule(lr){8-12}
        & & Time & Sem. & P & R & F1 & Time & Sem. & P & R & F1 & \\
        \midrule
        \multicolumn{13}{c}{\textit{Offline Models + Proactive Prompt}} \\
        \midrule
        Qwen2.5-VL-7B~\citep{qwen25vl}  & 7B 
        & 9.8 & 9.3 & 7.5 & 5.5 & 5.8
        & 5.1 & 2.9 & 9.0 & 2.0 & 3.0 & 4.4 \\

        Qwen3-VL-8B~\citep{qwen3vl}    & 8B 
        & 58.7 & 30.7 & 4.8 & 29.1 & 4.3
        & 7.3 & 4.0 & 2.9 & 3.3 & 1.8 & 3.1 \\

        \midrule
        \multicolumn{13}{c}{\textit{Open-Source Proactive Models}} \\
        \midrule
        VideoLLM-Online~\citep{videollmonline} & 8B 
        & 20.6 & 10.5 & 47.4 & 10.4 & \textbf{16.9}
        & 22.4 & 5.5 & 0.9 & 4.8 & 0.7 & \textbf{8.8} \\

        MMDuet~\citep{mmduet}          & 7B 
        & 30.2 & 16.1 & 2.1 & 9.3 & 3.4
        & 36.9 & 15.5 & 3.1 & 9.3 & 4.2 & 3.8 \\

        MMDuet2~\citep{mmduet2}         & 3B 
        & 40.2 & 15.4 & 6.4 & 12.4 & 6.5
        & 14.4 & 3.9 & 0.6 & 2.9 & 1.0 & 3.8 \\

        MiniCPM-o-4.5~\citep{minicpm_o_45}   & 9B 
        & 9.8 & 9.3 & 7.5 & 5.5 & 5.8
        & 5.1 & 2.9 & 9.0 & 2.0 & 3.0 & 4.4 \\

        Streamo~\citep{streamo}         & 3B 
        & 17.2 & 7.7 & 15.5 & 7.6 & 8.8
        & 1.1 & 0.3 & 0.0 & 0.3 & 0.0 & 4.4 \\

        \rowcolor{our_color!10}
        \textbf{StreamPro-SFT}  & \textbf{3B} 
        & 44.6 & 26.8 & 4.6 & 20.7 & 7.1
        & 39.4 & 13.9 & 0.9 & 9.7 & 1.5 & 4.3 \\
        \rowcolor{our_color!10}
        \textbf{StreamPro-GRPO} & \textbf{3B} 
        & 17.9 & 13.3 & 6.9 & 8.9 & 7.6 & 10.6 & 4.5 & 6.3 & 2.9 & 3.7 & 5.6 \\
        \rowcolor{our_color!20}
        \textbf{StreamPro-SFT}  & \textbf{4B} 
        & 37.2 & 25.6 & 6.8 & 20.9 & 8.8
        & 50.1 & 18.2 & 3.2 & 13.0 & \underline{4.3} & 6.5 \\
        \rowcolor{our_color!20}
        \textbf{StreamPro-GRPO} & \textbf{4B} 
        & 28.0 & 23.0 & 8.6 & 15.1 & \underline{10.6} & 13.6 & 8.2 & 8.9 & 5.6 & \textbf{6.3} & \underline{8.5} \\
        \bottomrule
    \end{tabularx}
\end{table*}
\begin{table*}[t]
\centering
\small
\renewcommand{\arraystretch}{1.2}
\setlength{\tabcolsep}{3.5pt}
\caption{Evaluation results on StreamingBench.}
\label{tab:streaming_bench_full}
\begin{adjustbox}{max width=\linewidth}
\begin{tabular}{l c ccccccccccc cccc c}
\toprule
\multirow{2}{*}{\textbf{Method}} & \multirow{2}{*}{\textbf{Param}} & \multicolumn{11}{c}{\textbf{Real-Time Visual Understanding}} & \multicolumn{4}{c}{\textbf{Contextual Understanding}} & \textbf{Overall} \\
\cmidrule(lr){3-13} \cmidrule(lr){14-17} \cmidrule(lr){18-18}
& & \textbf{OP} & \textbf{CR} & \textbf{CS} & \textbf{ATP} & \textbf{EU} & \textbf{TR} & \textbf{PR} & \textbf{SU} & \textbf{ACP} & \textbf{CT} & \textbf{Avg.} & \textbf{ACU} & \textbf{MCU} & \textbf{SQA} & \textbf{Avg.} & \textbf{Avg.} \\
\midrule
Human & - 
& 89.5 & 92.0 & 93.6 & 91.5 & 95.7 & 92.5 & 88.0 & 88.8 & 89.7 & 91.3 & 91.2 
& 88.8 & 90.4 & 95.0 & 91.4 & 91.3 \\
\midrule
\multicolumn{18}{c}{\textit{Offline Models}} \\
\midrule
Gemini 1.5 pro~\citep{gemini1.5} & - 
& 79.0 & 80.5 & 83.5 & 79.7 & 80.0 & \underline{84.7} & 77.8 & 64.2 & 72.0 & 48.7 & 75.8
& \underline{51.4} & 40.7 & \textbf{54.8} & \textbf{49.0} & 69.6 \\
GPT-4o~\citep{gpt4} & - 
& 77.1 & 80.5 & 83.9 & 76.5 & 70.2 & 83.8 & 66.7 & 62.2 & 69.1 & 49.2 & 73.3
& 41.2 & 38.4 & 32.8 & 37.5 & 65.1 \\
LLaVA-OneVision~\citep{llava-onevision} & 7B 
& 80.4 & 74.2 & 76.0 & 80.7 & 72.7 & 71.7 & 67.6 & 65.4 & 65.7 & 45.1 & 71.2
& 35.6 & 36.0 & 27.3 & 33.0 & 62.4 \\
Qwen2-VL-7B~\citep{qwen2vl} & 7B 
& 75.2 & 82.8 & 73.2 & 77.5 & 68.3 & 71.0 & 72.2 & 61.2 & 61.5 & 46.1 & 69.0
& 31.2 & 26.0 & 39.6 & 32.3 & 60.5 \\
InternVL2-8B~\citep{internvl3} & 8B 
& 68.1 & 60.9 & 69.4 & 77.1 & 67.7 & 62.9 & 59.3 & 53.3 & 55.0 & \underline{56.5} & 63.8
& 32.0 & 31.2 & 32.3 & 31.8 & 56.4 \\
\midrule
\multicolumn{18}{c}{\textit{Streaming Models}} \\
\midrule
VideoLLM-Online~\citep{videollmonline} & 8B 
& 39.1 & 40.1 & 34.5 & 31.1 & 46.0 & 32.4 & 31.5 & 34.2 & 42.5 & 27.9 & 36.0
& 24.2 & 29.2 & 30.8 & 28.1 & 34.1 \\
Flash-VStream~\citep{flash-vstream} & 7B 
& 25.9 & 43.6 & 24.9 & 23.9 & 27.3 & 13.1 & 18.5 & 25.2 & 23.9 & 48.7 & 25.9
& 24.8 & 25.2 & 26.8 & 25.6 & 25.8 \\
Dispider~\citep{dispider} & 7B 
& 74.9 & 75.5 & 74.1 & 73.1 & 74.4 & 59.9 & 76.1 & 62.9 & 62.2 & 45.8 & 67.5
& 39.6 & 27.7 & 34.8 & 34.0 & 59.8 \\
TimeChat-Online~\citep{timechat-online} & 7B & 80.2 & 82.0 & 79.5 & 83.3 & 76.1 & 78.5 & 78.7 & 64.6 & 69.6 & \textbf{58.0} & 75.3 & 39.2 & 31.6 & 41.6 & 37.5 & 66.6 \\
ViSpeak~\citep{vispeak} & 7B 
& 79.8 & \textbf{88.3} & 83.3 & 81.1 & 76.4 & 75.1 & 70.4 & 65.9 & 77.3 & 34.2 & 74.5 
& 38.8 & 36.8 & 44.0 & 39.9 & 66.5 \\
StreamForest~\citep{streamforest} & 7B & 83.1 & 82.8 & 82.7 & 84.3 & 77.5 & 78.2 & 76.9 & 69.1 & 75.6 & 54.4 & 77.3 & - & - & - & - & - \\
StreamAgent~\citep{streamagent} & 7B 
& 79.6 & 78.3 & 79.3 & 75.9 & 74.7 & 76.9 & \textbf{82.9} & 66.3 & 73.7 & 55.4 & 74.6
& 39.7 & 30.3 & 39.6 & 36.5 & 65.8 \\
QueryStream~\citep{querystream} & 7B & 82.4 & \underline{84.4} & 79.2 & 82.4 & 78.0 & 81.3 & 78.7 & 65.0 & 69.3 & 47.3 & 75.3 & - & - & - & - & - \\
Streamo-3B\textsuperscript{$\dagger$}~\citep{streamo} & 3B 
& 83.7 & 80.5 & 83.9 & \underline{84.6} & 80.5 & 81.0 & 71.3 & 67.5 & 71.6 & 37.2 & 75.8
& 43.6 & 38.4 & 41.2 & 41.1 & 67.8 \\

\rowcolor{our_color!10}
\textbf{StreamPro-SFT} & \textbf{3B} 
& 82.7 & 79.7 & 80.1 & \underline{84.6} & 81.1 & 82.2 & 76.9 & 69.9 & 76.7 & 44.7 & 77.1
& 42.4 & 34.8 & 42.8 & 40.0 & 68.5 \\

\rowcolor{our_color!10}
\textbf{StreamPro-GRPO} & \textbf{3B} 
& 81.6 & 79.7 & 81.1 & 81.7 & 80.5 & 82.6 & 75.9 & 69.9 & \underline{77.6} & 33.5 & 75.9
& 42.0 & 34.8 & 42.0 & 39.6 & 67.5 \\

\rowcolor{our_color!20}
\textbf{StreamPro-SFT} & \textbf{4B} 
& \textbf{86.7} & 78.9 & \textbf{88.0} & \textbf{84.9} & \underline{83.0} & \textbf{86.0} & \underline{82.4} & \textbf{74.0} & \textbf{78.7} & 33.0 & \underline{79.3}
& 48.8 & \textbf{43.6} & \underline{48.4} & \underline{46.9} & \underline{71.8} \\

\rowcolor{our_color!20}
\textbf{StreamPro-GRPO} & \textbf{4B} 
& \underline{85.4} & 77.3 & \underline{87.7} & 84.0 & \textbf{84.3} & 84.4 & 77.8 & \underline{73.2} & \textbf{78.7} & 50.0 & \textbf{79.8}
& \textbf{53.6} & \underline{43.2} & 43.2 & 46.7 & \textbf{72.1} \\

\bottomrule
\end{tabular}
\end{adjustbox}

\end{table*}

\begin{table*}[t]
\centering
\small
\setlength{\tabcolsep}{4pt}
\caption{Evaluation results on OVO-Bench.}
\label{tab:ovo_bench_full}
\begin{tabularx}{\linewidth}{l c *{7}{Y}}
\toprule
\multirow{2}{*}{\textbf{Method}} & \multirow{2}{*}{\textbf{Param}} &
\multicolumn{7}{c}{\textbf{Real-Time Visual Perception}} \\
\cmidrule(lr){3-9}
& & 
\textbf{OCR} & \textbf{ACR} & \textbf{ATR} & \textbf{STU} & \textbf{FPD} & \textbf{OJR} & \textbf{Avg.} \\
\midrule

Human & - & 94.0 & 92.6 & 94.8 & 92.7 & 91.1 & 94.0 & 93.2 \\
\midrule

\multicolumn{9}{c}{\textit{Offline Models}} \\
\midrule
Gemini 1.5 Pro~\citep{gemini1.5} & - & \textbf{85.9} & 67.0 & \textbf{79.3} & \textbf{58.4} & 63.4 & 62.0 & \underline{69.3} \\
GPT-4o~\citep{gpt4} & - & 69.8 & 64.2 & 71.6 & 51.1 & 70.3 & 59.8 & 64.5 \\
LLaVA-Video~\citep{llava-video} & 7B & 69.1 & 58.7 & 68.8 & 49.4 & \underline{74.3} & 59.8 & 63.4 \\
LLaVA-OneVision~\citep{llava-onevision} & 7B & 66.4 & 57.8 & 73.3 & 53.4 & 71.3 & 62.0 & 64.0 \\
Qwen2-VL-7B~\citep{qwen2vl} & 7B & 60.4 & 50.5 & 56.0 & 47.2 & 66.3 & 55.4 & 56.0 \\
InternVL2-8B~\citep{internvl3} & 8B & 67.1 & 60.6 & 63.8 & 46.1 & 68.3 & 56.5 & 60.4 \\
\midrule

\multicolumn{9}{c}{\textit{Streaming Models}} \\
\midrule
VideoLLM-online~\citep{videollmonline} & 8B & 8.1 & 23.9 & 12.1 & 14.0 & 45.5 & 21.2 & 20.8 \\
Flash-VStream~\citep{flash-vstream} & 7B & 24.2 & 29.4 & 28.5 & 33.7 & 25.7 & 28.8 & 28.4 \\
Dispider~\citep{dispider} & 7B & 57.7 & 49.5 & 62.1 & 44.9 & 61.4 & 51.6 & 54.5 \\
TimeChat-Online~\citep{timechat-online} & 7B & 75.2 & 46.8 & 70.7 & 47.8 & 69.3 & 61.4 & 61.9 \\
ViSpeak~\citep{vispeak} & 7B & 75.2 & 58.7 & 71.6 & 51.1 & \underline{74.3} & 66.9 & 66.3 \\
StreamForest~\citep{streamforest} & 7B & 68.5 & 53.2 & 71.6 & 47.8 & 65.4 & 60.9 & 61.2 \\
StreamAgent~\citep{streamagent} & 7B & 71.2 & 53.2 & 63.6 & 53.9 & 67.3 & 58.7 & 61.3 \\
QueryStream~\citep{querystream} & 7B & 75.2 & 49.5 & 69.8 & 50.0 & 71.3 & 62.5 & 63.1 \\
Thinking-QwenVL~\citep{thinking-qwenvl} & 7B & 74.1 & 57.2 & 68.1 & \underline{55.3} & \textbf{75.0} & 58.3 & 64.7 \\
Streamo-3B\textsuperscript{$\dagger$}~\citep{streamo} & 3B & 73.8 & 51.4 & 74.1 & 47.2 & 57.4 & 63.0 & 61.2 \\
Streamo-7B~\citep{streamo} & 7B & 79.2 & 57.8 & \underline{75.0} & 49.4 & 64.4 & 70.1 & 66.0 \\
\rowcolor{our_color!10}
\textbf{StreamPro-SFT}  & \textbf{3B} & 73.2 & 51.4 & \underline{75.0} & 50.6 & 58.4 & 64.1 & 62.1 \\
\rowcolor{our_color!10}
\textbf{StreamPro-GRPO} & \textbf{3B} & 74.5 & 59.6 & 74.1 & 50.6 & 59.4 & 66.8 & 64.2 \\
\rowcolor{our_color!20}
\textbf{StreamPro-SFT}  & \textbf{4B} & 79.2 & \textbf{68.8} & 69.0 & 53.9 & 66.3 & \textbf{71.7} & 68.2 \\
\rowcolor{our_color!20}
\textbf{StreamPro-GRPO} & \textbf{4B} & \underline{82.6} & \underline{67.9} & 73.3 & 55.1 & 68.3 & \underline{71.2} & \textbf{69.7} \\
\bottomrule
\end{tabularx}
\end{table*}

\section{StreamPro-Bench Samples}
\label{sec:streampro_bench_samples}

\begin{figure*}[p]
    \centering
    \includegraphics[width=0.9\linewidth]{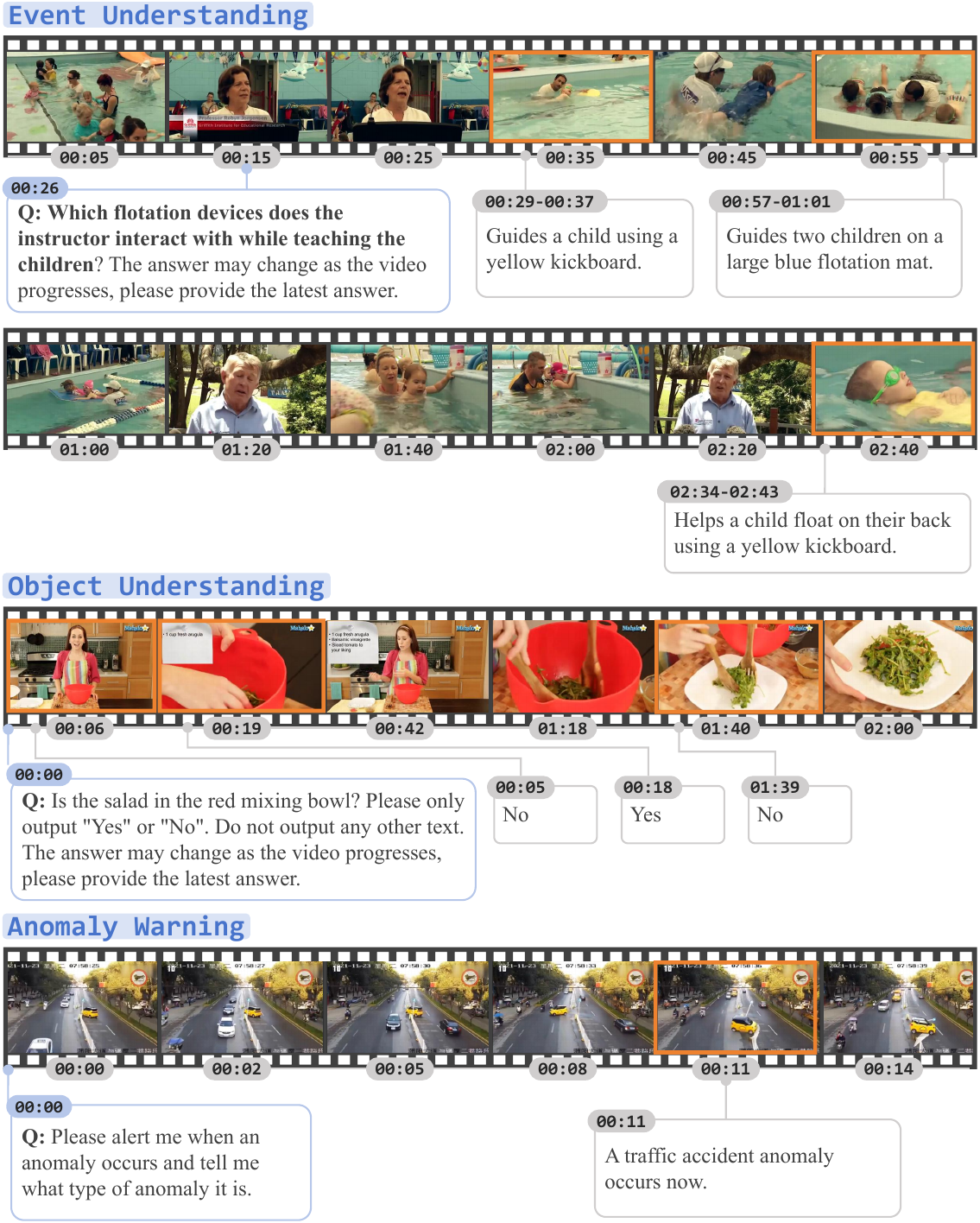}
    \caption{Benchmark samples of Perceptual Understanding tasks}
    \label{fig:perceptual_understanding_bench_case}
\end{figure*}

\begin{figure*}[p]
    \centering
    \includegraphics[width=0.9\linewidth]{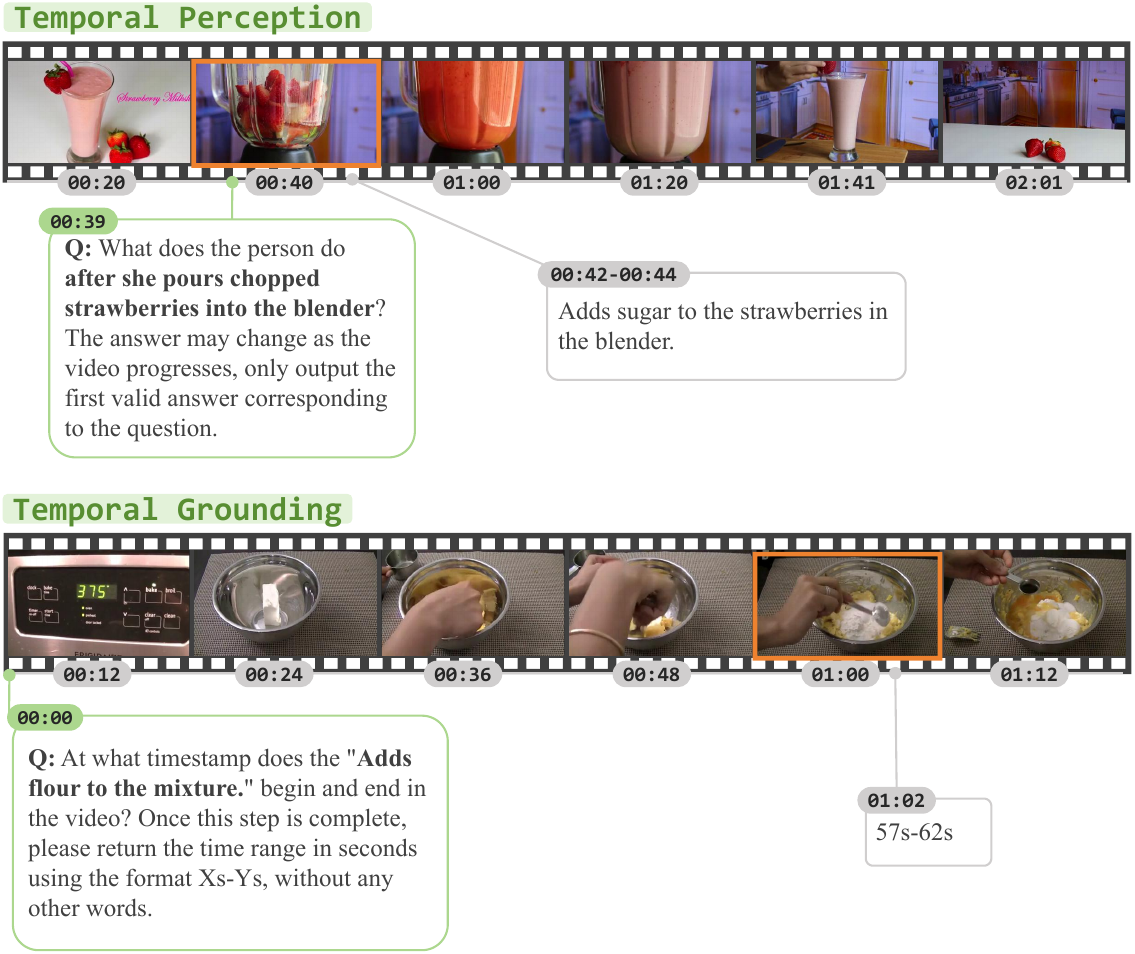}
    \caption{Benchmark samples of Temporal Reasoning tasks}
    \label{fig:temporal_reasoning_bench_case}
\end{figure*}

\begin{figure*}[p]
    \centering
    \includegraphics[width=0.9\linewidth]{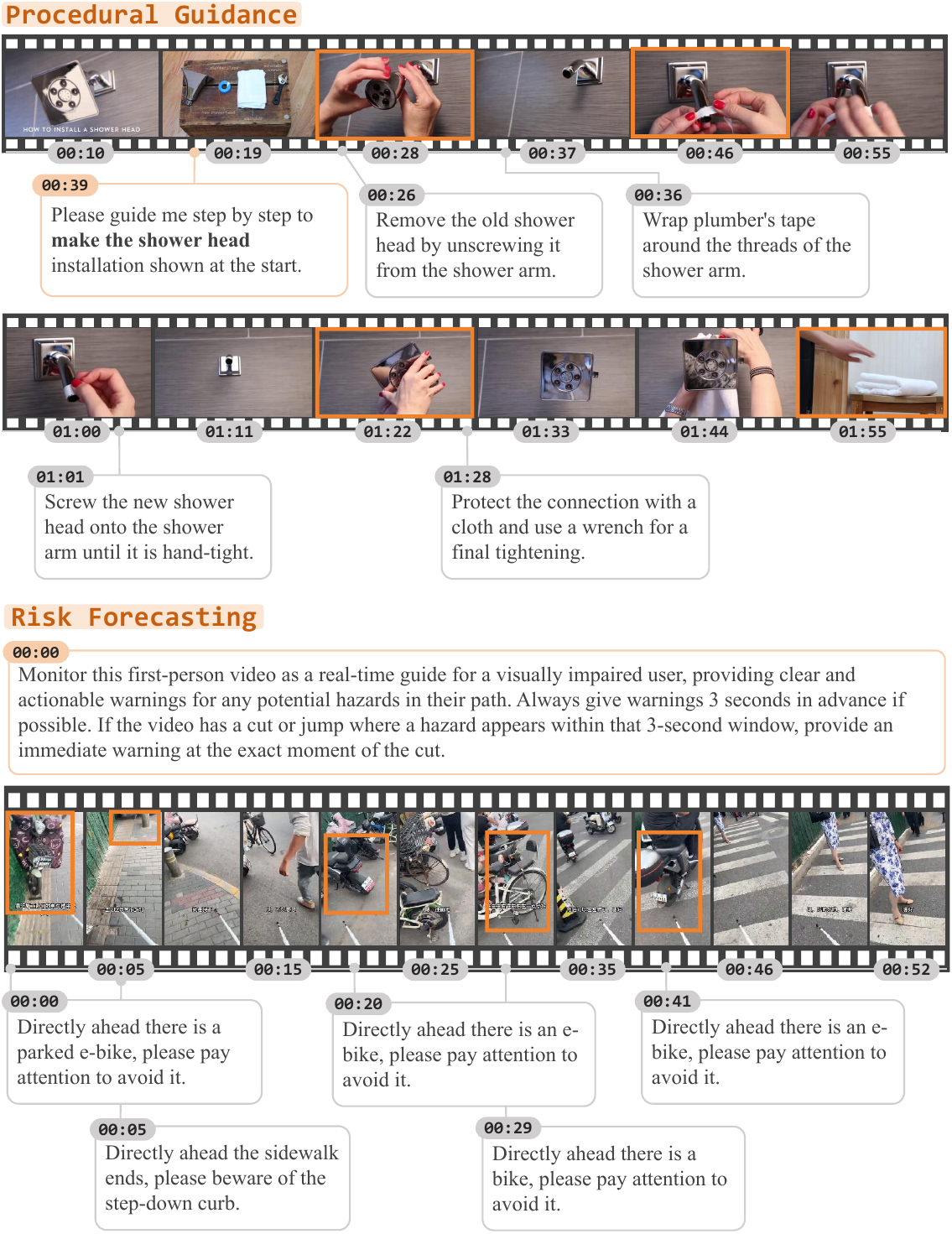}
    \caption{Benchmark samples of Anticipatory Assistance tasks.}
    \label{fig:anticipatory_assistance_bench_case}
\end{figure*}

We present additional samples of the benchmark tasks, as shown in 
Figure~\ref{fig:temporal_reasoning_bench_case}, Figure~\ref{fig:perceptual_understanding_bench_case}, and Figure~\ref{fig:anticipatory_assistance_bench_case}.

\section{Case Study}
\label{sec:case_study}

\begin{figure*}[t]
    \centering
    \includegraphics[width=0.8\linewidth]{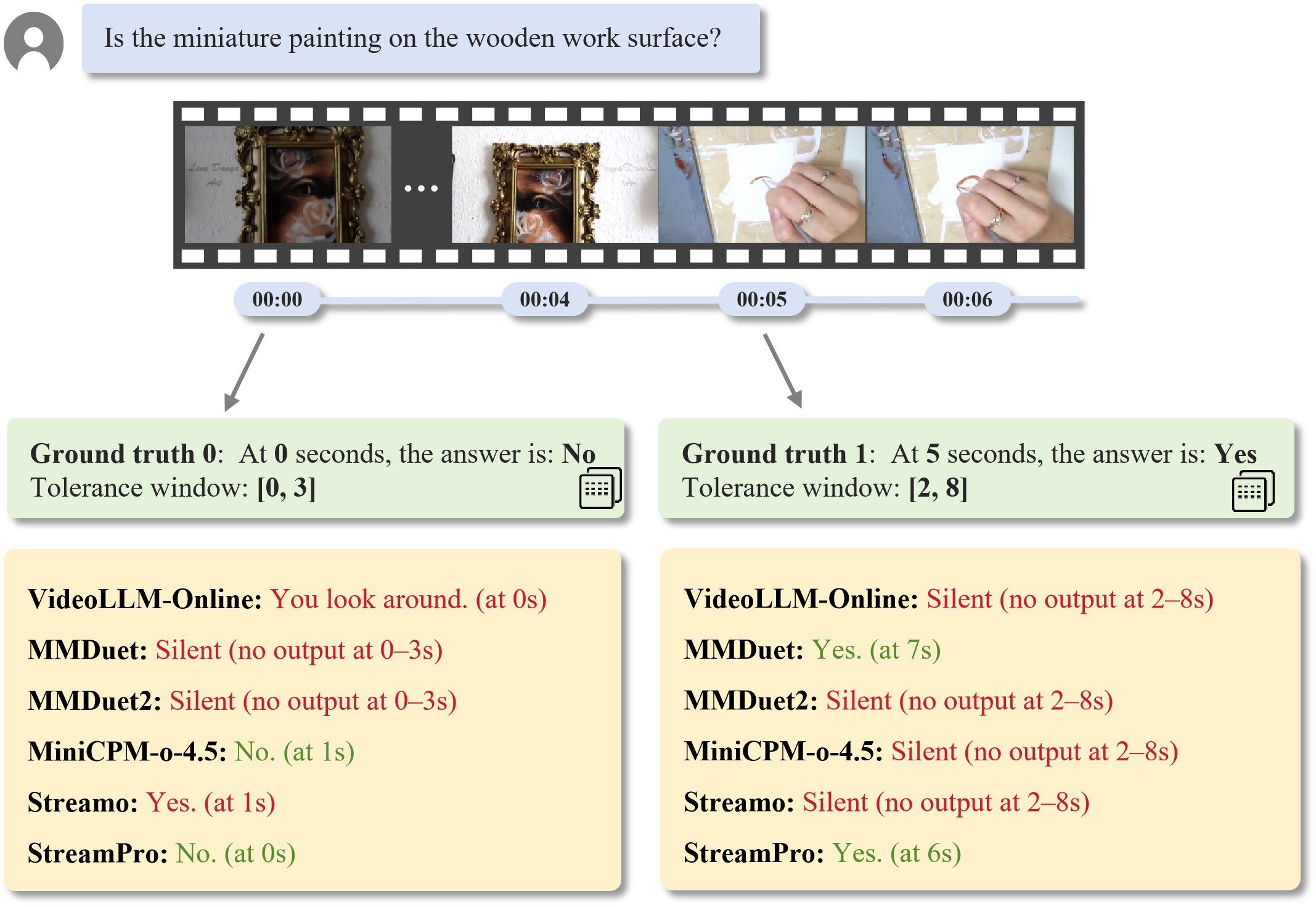}
    \caption{Comparison of different models on object understanding tasks.}
    \label{fig:ou_case_1}
\end{figure*}

\begin{figure*}[t]
    \centering
    \includegraphics[width=0.8\linewidth]{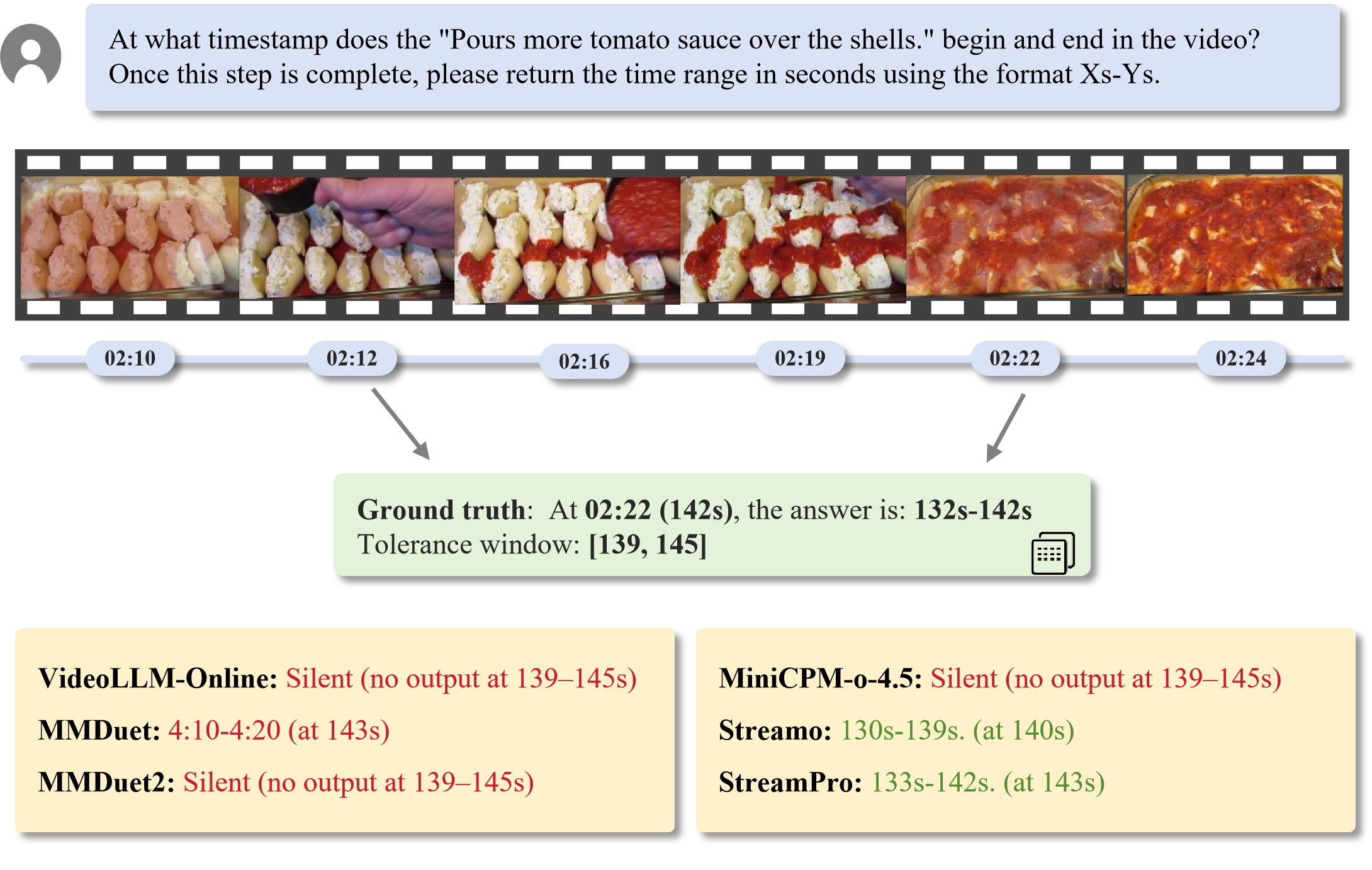}
    \caption{Comparison of different models on temporal grounding tasks.}
    \label{fig:tg_case_1}
\end{figure*}

\begin{figure*}[t]
    \centering
    \includegraphics[width=\linewidth]{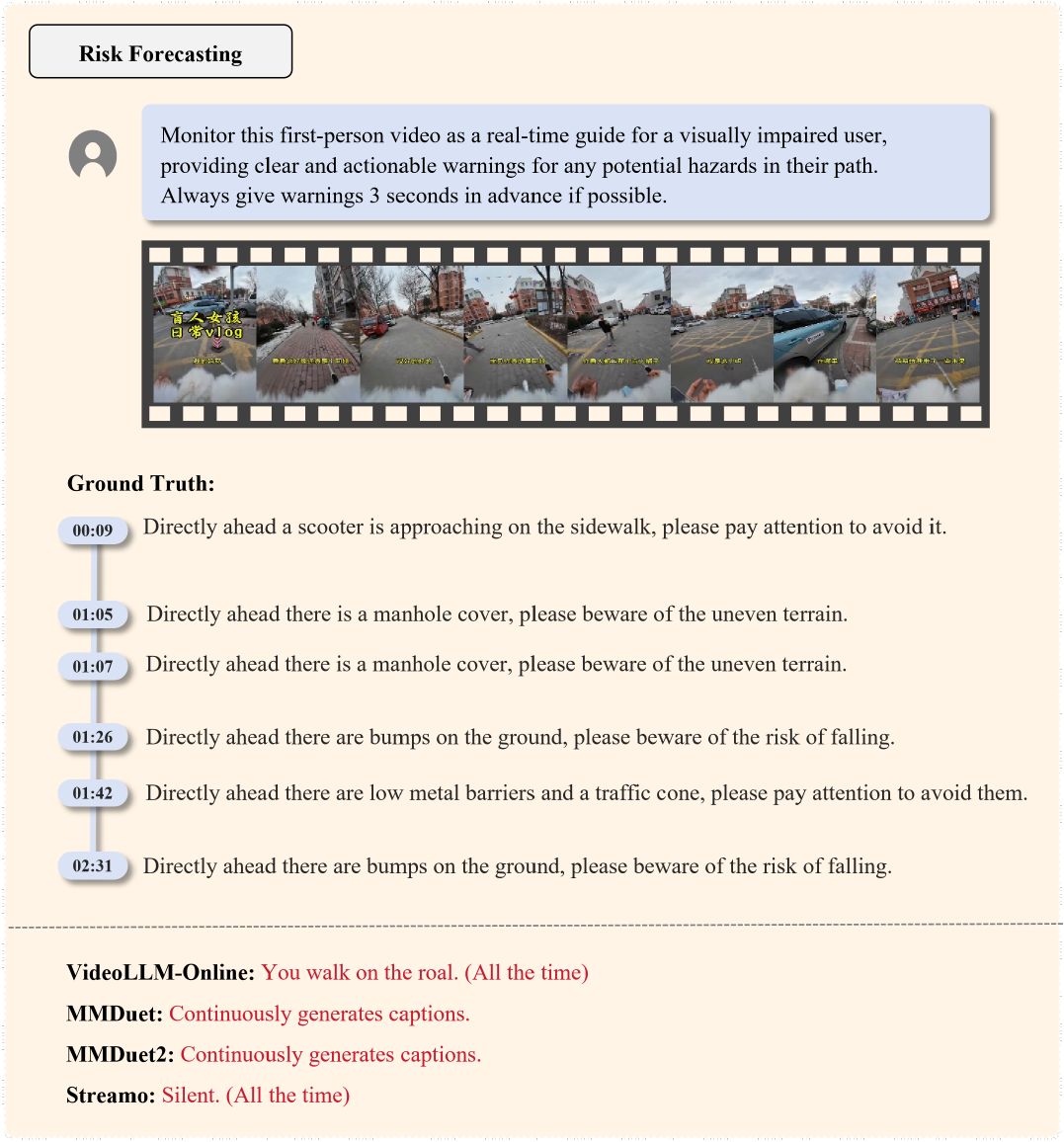}
    \caption{Comparison of different models on risk forecasting tasks.}
    \label{fig:rk_case_1}
\end{figure*}

\begin{figure*}[t]
    \centering
    \includegraphics[width=\linewidth]{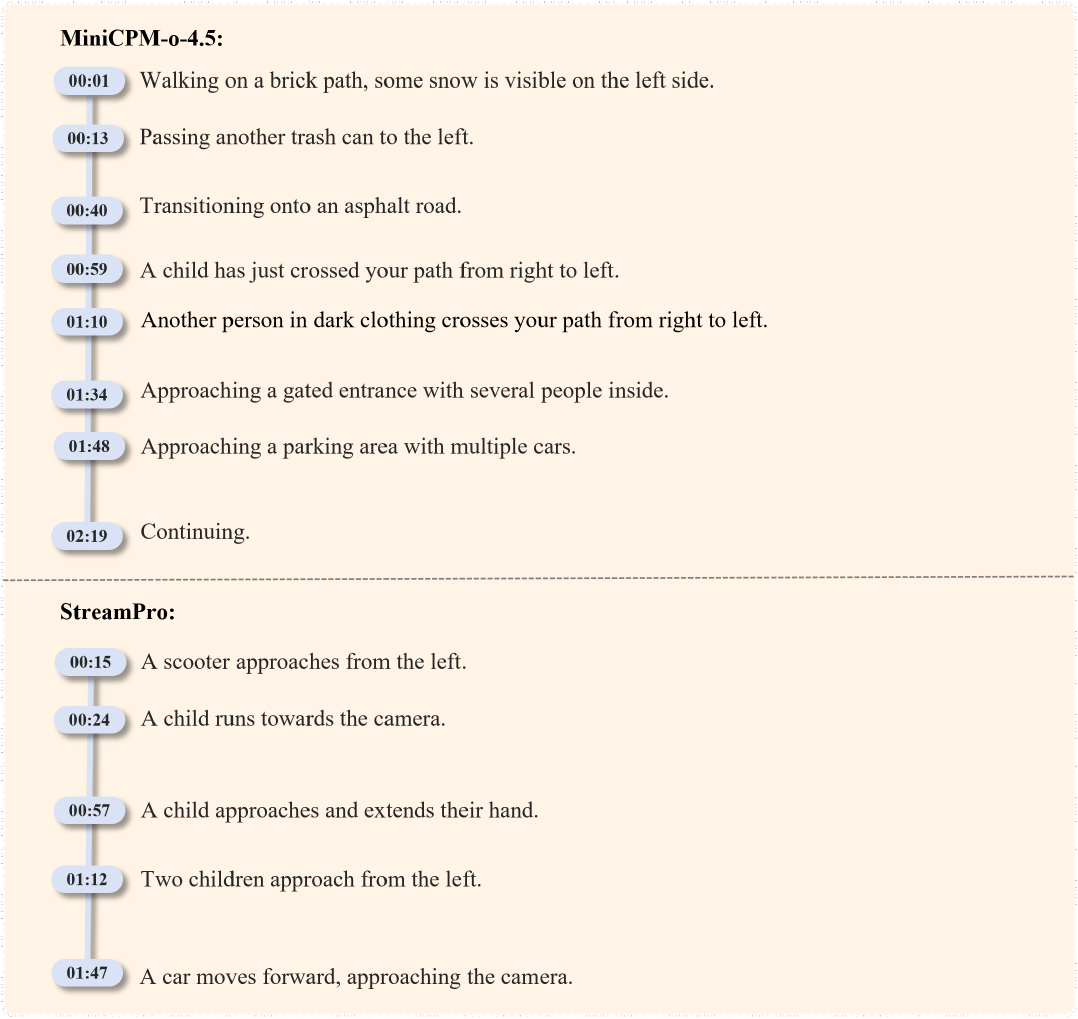}
    \caption{Comparison of different models on risk forecasting tasks.}
    \label{fig:rk_case_2}
\end{figure*}

We compare the outputs of different models to comprehensively evaluate the proactive capabilities of StreamPro, as shown in Figure~\ref{fig:ou_case_1}, \ref{fig:tg_case_1}, \ref{fig:rk_case_1}, and \ref{fig:rk_case_2}.
StreamPro demonstrates strong performance on both object understanding and temporal grounding tasks, consistently producing accurate and well-timed responses.
In contrast, for the risk forecasting examples, all existing models still struggle to achieve satisfactory results, reflecting the limited anticipatory assistance capability in this more challenging setting.

\section{Prompt Templates}
\label{sec:prompt_templates}

Here we present the system prompt used in StreamPro framework, the prompt used for evaluating offline models, the prompt for rubric generation, and the prompt used for open-ended evaluation.

\begin{figure*}[p]
\begin{PromptBox}{System Prompt}
You are a helpful assistant specializing in streaming video analysis. You will receive input frame by frame, where each frame is labeled with absolute time intervals in the exact format \texttt{<Xs-Ys>} (e.g., \texttt{<0s-1s>}). Follow the rules below strictly:

 \vspace{0.8em} 
\textbf{Response Protocol:}
\begin{itemize}[leftmargin=1.5em, nosep]
    \item Use \texttt{</Silence>} when:
        \begin{itemize}[nosep]
            \item A relevant event has not yet fully completed, \textbf{or}
            \item The current input is irrelevant to the given question.
        \end{itemize}
            
    \item Use \texttt{</Response>} only when:
        \begin{itemize}[nosep]
            \item An event has fully concluded, \textbf{or}
            \item The available information is sufficient to fully answer the question.
        \end{itemize}
        At this point, provide a complete and final answer.
\end{itemize}

 \vspace{0.8em} 
\textbf{Additional Constraints:}
\begin{itemize}[leftmargin=1.5em, nosep]
    \item Do not provide partial answers.
    \item Do not speculate beyond the given information.
    \item Every valid answer must begin with \texttt{</Response>}.
\end{itemize}
\end{PromptBox}

\vspace{3em}

\begin{PromptBox}{Offline Model Evaluation Prompt}
You are a Streaming Video Analysis Expert.
You will receive a video sequence frame-by-frame.
A user Query will be issued once at an arbitrary time during the video stream. Before the Query is issued, observe the video carefully. After the Query is issued, follow the decision logic below.

\vspace{0.8em}
\textbf{Objective:}
Determine whether the visual information observed so far (from the beginning of the video up to the current moment) is sufficient to answer the user's Query.

\vspace{0.8em}
\textbf{Decision Logic:}
For each input frame sequence after the Query has been issued:
\begin{enumerate}[leftmargin=1.5em, nosep]
    \item Is the visual information observed so far sufficient to answer the Query without guessing?
    \begin{itemize}[nosep]
        \item If NO $\rightarrow$ reply \texttt{Wait}.
        \item If YES $\rightarrow$ proceed to Step 2.
    \end{itemize}

    \item Does the inferred answer differ from the last answer you provided?
    \begin{itemize}[nosep]
        \item If NO $\rightarrow$ reply \texttt{Wait}.
        \item If YES (First Trigger) $\rightarrow$ reply with the actual answer.
        \item If YES (Answer Update due to new evidence) $\rightarrow$ reply with the updated answer.
    \end{itemize}
\end{enumerate}

\vspace{0.8em}
\textbf{Output Constraint:}
\begin{itemize}[leftmargin=1.5em, nosep]
    \item Do not output anything other than \texttt{Wait} or the actual content of the answer.
\end{itemize}
\end{PromptBox}
\end{figure*}

\begin{figure*}[p]
\begin{PromptBox}{Rubric Generation Prompt}
You are designing evaluation rubrics for a streaming video QA system.
The rubrics will be used to evaluate model trajectories AFTER generation — the model's trajectory is not available yet.
You see the ground truth and the video.

\vspace{0.8em}
\textbf{Input:}
\begin{itemize}[leftmargin=1.5em, nosep]
    \item \textbf{Question:} \{question\}
    \item \textbf{Ground Truth Answer(s):} \{ground\_truth\_text\}
\end{itemize}

\vspace{0.8em}
\textbf{Task:}
Design 6--10 binary (pass/fail) evaluation rubrics that assess a model's streaming trajectory quality.
Each rubric must be \textbf{SPECIFIC} to this particular video and question — not generic criteria.

\vspace{0.8em}
\textbf{Important Requirements:}
\begin{itemize}[leftmargin=1.5em, nosep]
    \item Each rubric must be an \textbf{ASSERTIVE statement}:
    \begin{itemize}[nosep]
        \item Use words like \texttt{must}, \texttt{shall not}, \texttt{is prohibited}.
        \item Do NOT write vague descriptions (e.g., ``the model should do X well'').
        \item Write concrete assertions (e.g., ``The response must include Y — missing this is not allowed'').
    \end{itemize}
\end{itemize}

\vspace{0.8em}
\textbf{Rubric Dimensions (each rubric covers one):}
\begin{itemize}[leftmargin=1.5em, nosep]
    \item \textbf{Granularity:}
    \begin{itemize}[nosep]
        \item Assess whether step decomposition matches the natural structure of the video.
        \item The model shall not over-split atomic actions.
        \item The model shall not merge distinct actions into vague descriptions.
    \end{itemize}

    \item \textbf{Sequencing:}
    \begin{itemize}[nosep]
        \item Steps must follow correct temporal and physical logic.
        \item The model shall not skip necessary intermediate steps.
    \end{itemize}

    \item \textbf{Coverage:}
    \begin{itemize}[nosep]
        \item All essential actions or phases must be included.
        \item Missing key stages is not allowed.
    \end{itemize}
\end{itemize}

\vspace{0.8em}
\textbf{Post-check (self-review):}
\begin{itemize}[leftmargin=1.5em, nosep]
    \item Each rubric must tightly match the specific video and QA pair.
    \item No rubric should be generic or broadly applicable.
    \item All three dimensions must be covered.
    \item Total number must be between 6 and 10.
    \item Each rubric must be written as a concrete assertion.
\end{itemize}

\vspace{0.8em}
\textbf{Output Format:}
Return \textbf{ONLY} the final JSON array:
\begin{verbatim}
[
  {
    "rubric_id": 0,
    "dimension": "granularity|sequencing|coverage",
    "rubric": "Assertive statement about what to check"
  }
]
\end{verbatim}

\end{PromptBox}
\end{figure*}

\begin{figure*}[p]
\begin{PromptBox}{StreamPro-Bench Open-Ended Evaluation Prompt}
You are an expert evaluator judging whether a model's answer provides a reasonable and factually correct response that directly addresses the question, based on the reference answer.

\vspace{0.8em}
\textbf{Evaluation Guideline:}
\begin{itemize}[leftmargin=1.5em, nosep]
    \item Focus on whether the model provides information that directly answers the question.
    \item The answer does not need to reproduce all details from the reference—it only needs to provide factually correct and relevant information.
    \item An answer that captures the essential information required by the question should be considered strong, even if it omits descriptive details.
    \item Accept simplified, rephrased, or high-level responses as long as they are consistent with the reference and do not contradict known facts.
    \item Do not deduct points for omitting secondary or illustrative details when the key information needed to answer the question is present, or for using concise or abstract phrasing.
    \item Only penalize if the answer is factually wrong, fails to provide relevant information, or is so vague that it does not actually answer the question.
\end{itemize}

\vspace{0.8em}
\textbf{Scoring (integer 0--5):}
\begin{itemize}[leftmargin=1.5em, nosep]
    \item 5: Fully accurate and complete answer.
    \item 4: Correct and sufficiently informative answer; may omit non-essential details but contains the key information.
    \item 3: Partially relevant but missing some important information.
    \item 2: Tangential or speculative without solid factual grounding.
    \item 1: Factually incorrect.
    \item 0: No attempt to answer or completely off-topic.
\end{itemize}

\vspace{0.8em}
\textbf{Output Format:}
\begin{itemize}[leftmargin=1.5em, nosep]
    \item Return a valid JSON object with exactly two keys:
    \begin{itemize}[nosep]
        \item \texttt{"explanation"}: one sentence focusing on whether the answer provides correct and relevant information for the question
        \item \texttt{"score"}: an integer from 0 to 5
    \end{itemize}
    \item Output only the JSON. No other text, markdown, or commentary.
\end{itemize}

\vspace{0.8em}
\textbf{Inputs:}
\begin{itemize}[leftmargin=1.5em, nosep]
    \item Question: \{question\}
    \item Predicted Answer: \{model\_output\}
    \item Correct Answer: \{reference\_answer\}
\end{itemize}

\end{PromptBox}
\end{figure*}

\end{document}